%% file: eccv2020submissionCR.tex
\documentclass[runningheads]{llncs}
\usepackage{graphicx}
\usepackage{tikz}
\usepackage{comment} 
\usepackage{amsmath,amssymb} % define this before the line numbering.
\usepackage{color}
\usepackage{cite} %added by Miao

\begin{document}
% \renewcommand\thelinenumber{\color[rgb]{0.2,0.5,0.8}\normalfont\sffamily\scriptsize\arabic{linenumber}\color[rgb]{0,0,0}}
% \renewcommand\makeLineNumber {\hss\thelinenumber\ \hspace{6mm} \rlap{\hskip\textwidth\ \hspace{6.5mm}\thelinenumber}}
% \linenumbers
\pagestyle{headings}
\mainmatter
\def\ECCVSubNumber{100}  % Insert your submission number here

\title{DVI: Depth Guided Video Inpainting for Autonomous Driving} % Replace with your title

% INITIAL SUBMISSION 
\begin{comment}
\titlerunning{ECCV-20 submission ID \ECCVSubNumber} 
\authorrunning{ECCV-20 submission ID \ECCVSubNumber} 
\author{Anonymous ECCV submission}
\institute{Paper ID \ECCVSubNumber}
\end{comment}
%******************

% CAMERA READY SUBMISSION
%\begin{comment}
\titlerunning{Depth Guided Video Inpainting}
% If the paper title is too long for the running head, you can set
% an abbreviated paper title here
%
\author{Miao Liao \and
Feixiang Lu \and
Dingfu Zhou \and
Sibo Zhang \and
Wei Li \and
Ruigang Yang}
\authorrunning{M. Liao et al.}
% First names are abbreviated in the running head.
% If there are more than two authors, 'et al.' is used.
%
\institute{Baidu Research, Baidu Inc \\
\email{miao.liao@gmail.com, lufeixiang@baidu.com,  \{dingfuzhou, sibozhang1, liweimcc\}@gmail.com, ryang2@uky.edu}
}
%\end{comment}
%******************
\maketitle

\begin{abstract}
To get clear street-view and photo-realistic simulation in autonomous driving, we present an automatic video inpainting algorithm that can remove traffic agents from videos and synthesize missing regions with the guidance of depth/point cloud. By building a dense 3D map from stitched point clouds, frames within a video are geometrically correlated via this common 3D map. In order to fill a target inpainting area in a frame, it is straightforward to transform pixels from other frames into the current one with correct occlusion. Furthermore, we are able to fuse multiple videos through 3D point cloud registration, making it possible to inpaint a target video with multiple source videos. The motivation is to solve the long-time occlusion problem where an occluded area has never been visible in the entire video. To our knowledge, we are the first to fuse multiple videos for video inpainting. To verify the effectiveness of our approach, we build a large inpainting dataset in the real urban road environment with synchronized images and Lidar data including many challenge scenes, e.g., long time occlusion. The experimental results show that the proposed approach outperforms the state-of-the-art approaches for all the criteria, especially the RMSE (Root Mean Squared Error) has been reduced by about $\textbf{13}\%$.
\end{abstract}

\input{1-intro}

\input{2-related}

\input{3-method}

\input{4-exp}

\input{5-con}

\clearpage
% ---- Bibliography ----
%
% BibTeX users should specify bibliography style 'splncs04'.
% References will then be sorted and formatted in the correct style.
%
\bibliographystyle{splncs04}
\bibliography{egbib}
\end{document}

%% file: 1-intro.tex
\section{Introduction}
As computational power increases, multi-modality sensing has become more and more popular in recent years. Especially in the area of Autonomous Driving (AD), multiple sensors are combined to overcome the drawbacks of individual ones, which can provide redundancy for safety. Nowadays, most self-driving cars are equipped with lidar and cameras for both perception and mapping. Simulation systems have become essential to the development and validation of AD  technologies. Instead of using computer graphics to create virtual driving scenarios, Li et al.~\cite{Lieaaw0863} proposed to augment real-world pictures with simulated traffic flow to create photorealistic simulation images and renderings. One key component in their pipeline is to remove those moving agents on the road to generate clean background street images. AutoRemover~\cite{zhang2019autoremover} generated those kinds of data using the augmented platform and proposed a video inpainting method based on the deep learning techniques. Those map service companies, which display street-level panoramic views in their map Apps, also choose to place depth sensors in addition to image sensors on their capture vehicles. Due to privacy protection, those street view images have to be post-processed to blur human faces and vehicle license plates before posted for public access. There is a strong desire to totally remove those agents on the road for better privacy protection and more clear street images.

Significant progress has been made in image inpainting in recent years. The mainstream approaches~\cite{simakov2008summarizing, Darabi12:ImageMelding12, Efros:2001:IQT:383259.383296} adopt the patch-based method to complete missing regions by sampling and pasting similar patches from known regions or other source images. The method has been naturally extended to video inpainting, where not only spatial coherence but also temporal coherence are preserved.

The basic idea behind video inpainting is that the missing regions/pixels within a frame are observed in some other frames of the same video. Under this observation, some prior works~\cite{Xu_2019_CVPR, Huang-SigAsia-2016, wang2018videoinp} use optical flow as guidance to fill the missing pixels either explicitly or implicitly. They are successfully applied in different scenarios with seamless inpainting results. However, flow computation suffers from textureless areas, no matter it's learning based or not. Furthermore, perspective changes in the video could also degrade the quality of optical flow estimation. These frame-wise flow errors are accumulated when we fill missing pixels from a temporally distant frame, resulting in distorted inpainting results, which will be shown in the experiment section. 
 
The emergence of deep learning, especially Generative Adversarial Networks (GAN), has provided us a powerful tool for inpainting. For images, \cite{IizukaSIGGRAPH2017, pathakCVPR16context, yu2018generative} formulate inpainting as
a conditional image generation problem. Although formulated differently, GAN based inpainting approaches are essentially the same as the patch-based approach, since the spirit is still looking for similar textures in the training data and fill the holes. Therefore, they have to delicately choose their training data to match the domain of the input images. And domain adaptation is not an easy task once the input images come from different scenarios. Moreover, GAN-based approaches share the same problem as the patch-based methods that they are poor at handling perspective changes in images.

As image+depth sensors become standard for AD cars, we propose a method to inpaint street-view videos with the guidance of depth. Depending on the tasks, target objects are either manually labeled or automatically detected in color images, and then removed from their depth counterpart. A 3D map is built by stitching all point clouds together and projected back onto individual frames. Most of the frame pixels are assigned with a depth value via 3D projection and those remaining pixels get their depth by interpolation. With a dense depth map and known extrinsic camera parameters, we are able to sample colors from other frames to fill holes within the current frame.  These colors serve as an initial guess for those missing pixels, followed by regularization enforcing spatial and photometric smoothness. After that, we apply color harmonization to make smooth and seamless blending boundaries. In the end, a moving average is applied along the optical flow to make the final inpainted video look smooth temporally.   

Unlike learning-based methods, our approach can’t inpaint occluded areas if they are never visible in the video. To solve this problem, we propose fusion inpainting, which makes use of multiple video clips to inpainting a target region. Compared to state-of-the-art inpainting approaches, we are able to preserve better details in the missing region with correct perspective distortion. Temporal coherence is implicitly enforced since the 3D map is consistent across all frames. We are even able to inpaint multiple video clips captured at different times by registering all the frames into a common 3D point map. Although our experiments are conducted on datasets captured from a self-driving car, the proposed method is not limited to this scenario only. It can be generalized to both indoor and outdoor scenarios, as long as we have synchronized image+depth data.

In this paper, we propose a novel video inpainting method with the guidance of 3D maps in AD scenarios. We avoid using deep learning-based methods so that our entire pipeline only runs on CPUs. This makes it easy to be generalized to different platforms and different use cases because it doesn't require GPUs and domain adaptation of training data. 3D map guided inpainting is a new direction for the inpainting community to explore, given that more and more videos are accompanied with depth data. The main contributions of this paper are listed as follows:

% \textbf{Depth Guided Algorithm} We are the first to propose depth guided inpainting.

%\textbf{Depth Guided Algorithm} 
\begin{enumerate}
\item {
We propose a novel approach of depth guided video inpainting for autonomous driving;}

%\textbf{Fusion of Multiple Videos} 
\item {We are the first to fuse multiple videos for inpainting, in order to solve the long time occlusion problem;}
% \textbf{Color Harmonization} We address the issue of color harmonization in video inpainting.
% \textbf{Candidate Color Criteria} We propose criteria to select the best candidate colors to maximize texture preservation and minimize perspective distortion.
%\textbf{Dataset} 
\item {We collect a new dataset in the urban road with synchronized images and Lidar data including many challenge inpainting scenes such as long time occlusion;}

\item {Furthermore, we designed Candidate Color Sampling Criteria and Color Harmonization for inpainting. Our approach shows smaller RMSE compared with other state-of-art methods. }
\end{enumerate}

%% file: 2-related.tex
\section{Related Work}

The principle of inpainting is essentially filling the target holes by borrowing appearance information from known sources. The sources could be regions other than the hole in the same image, images from the same video or images/videos of similar scenarios. It's critical to reduce the search space for the right pixels. Following different cues, prior works can be categorized into 3 major classes: propagation-based inpainting, patch-based inpainting, and learning-based inpainting. 

\textbf{Propogation-based Inpainting}. Propagation-based  methods~\cite{Ballester:2001:FJI:2318999.2320160, 2015ITIP243034E} extrapolate boundary pixels around the holes for image completion.  These approaches are successfully applied to regions of uniform colors. However, it has difficulties to fill large holes with rich texture variations. Thus, Propagation-based approaches usually repair small holes and scratches in an image. 

\textbf{Patch-based Inpainting}. Patch-based methods~\cite{Bertalmio:2003:SST:2319023.2320434, Darabi12:ImageMelding12, Efros:2001:IQT:383259.383296, simakov2008summarizing} on the other hand, not only look at the boundary pixels but also search in the other regions/images for similar appearance in order to complete missing regions. This kind of approach has been extended to the temporal domain for video inpainting~\cite{Newson:2013:TFG:2534008.2534019, vics, Shih:2009:EVI:1641626.1641630}.   Huang et al.~\cite{Huang-SigAsia-2016} jointly estimate optical flow and color in the missing regions to address the temporal consistency problem. In general, patch-based methods can better handle non-stationary visual data. As suggested by its name, Patch-based methods depend on reliable pixel matches to copy and paste image patches to missing regions. When a pixel match can't be robustly obtained, for example in cases of big perspective changes or illumination changes, the inpainting results are problematic. 

\textbf{Learning-based Inpainting}. The success of deep learning techniques inspires recent works on applying it for image inpainting.   Ren et al.~\cite{NIPS2015_5774} adds a few feature maps in the new Shepard layers, achieving stronger results than a much deeper network architecture. Generative Adversarial Networks(GAN~\cite{NIPS2014_5423}) was first introduced to generate novel photos. It's straightforward to extend it to inpainting by formulating inpainting as a conditional image generation problem~\cite{IizukaSIGGRAPH2017, pathakCVPR16context, yu2018generative}. Pathak et al.~\cite{pathakCVPR16context} proposed context encoders, which is a convolutional neural network trained to generate the contents of an arbitrary image region conditioned on its surroundings. The context encoders are trained to both understand the content of the entire image, and produce a plausible hypothesis for the missing parts. Iizuka et al.~\cite{IizukaSIGGRAPH2017} used global and local context discriminators to distinguish real images from fake ones. The global discriminator looks at the entire image to ensure it is coherent as a whole, while the local discriminator looks only at a small area centered at the completed region to ensure the local consistency of the generated patches. More recently, Yu et al.~\cite{yu2018generative} presented a contextual attention mechanism in a generative inpainting framework, which further improves the inpainting quality. For video inpainting, Xu et al.~\cite{Xu_2019_CVPR} formulated an effective framework that is specially designed to exploit redundant information across video frames. They first synthesize a spatially and temporally coherent optical flow field across video frames, then the synthesized flow field is used to guide the propagation of pixels to fill up the missing regions in the video.

%% file: 3-method.tex
\section{Proposed Approach}

\begin{figure*}[t!]
\centering
\includegraphics[width=1.0\linewidth]{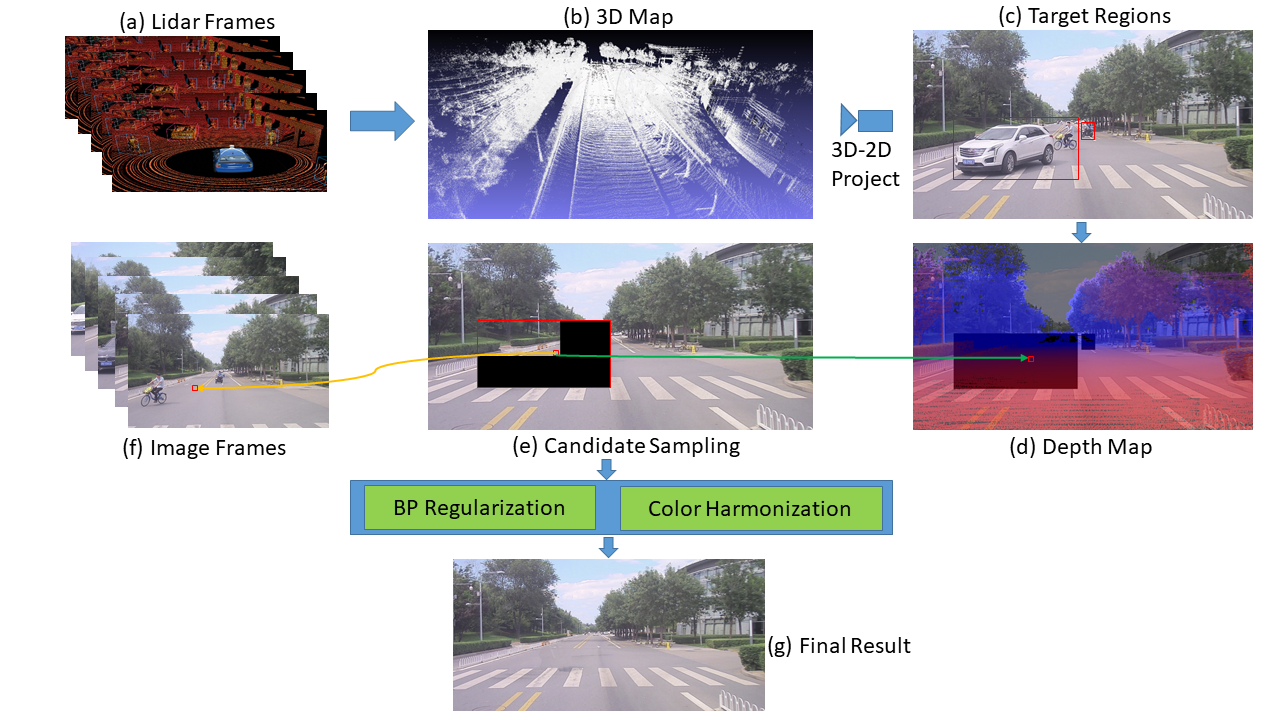}
   \caption{Frame-wise point clouds (a) are stitched into a 3D map (b) using LOAM. The 3D map is projected onto a frame (c) to generate a depth map. For each pixel in the target region (e), we use its depth (d) as guidance to sample colors from other frames (f). Final pixel values are determined by BP regularization and color harmonization to ensure photometric consistency. (g) shows the final inpainting result.}
\label{fig:pipeline}
\end{figure*}

Fig.~\ref{fig:pipeline} shows a brief pipeline of our approach. A 3D map is first built by stitching all point clouds together, and projected back onto individual frames. With dense depth map and known extrinsic camera parameters, we are able to sample candidate colors from other frames to fill holes within current frame. Then, a belief propagation based regularization is applied to make sure pixel colors within the inpainting region are consistent with each other. It is followed by a color harmonization step which ensures that colors within inpainting region are consistent with outside regions. More details will be described in the following subsections.

%A 3D map is first built by stitching all point cloud together and projected back onto individual frames. With dense depth maps and known extrinsic camera parameters, we are able to sample colors from other frames to fill holes within the current frame. More details will be described in the following subsections. Figure~\ref{fig:pipeline} shows a brief pipeline of our approach. 

\subsection{3D Depth Map}

\textbf{Dynamic Object Removal}. We first remove the moving objects from the point cloud, only keep the background points in the final 3D map. It is straight-forward to do so once the calibration between the depth sensor and the image sensor is performed. All points that are projected in the bounding box of the image are removed. The bounding boxes can be automatically detected or manually labeled. Alternatively, we can use PointNet++~\cite{qi2017pointnetplusplus} to detect and remove those typical moving objects directly from the point cloud.

\textbf{3D Map Stitching}. For lidar sensors, LOAM~\cite{Zhang-2014-7903} is a quite robust tool to fuse the multiple frames to build the 3D map. It is capable to match and track geometric features even with a sparse 16-beam lidar. For other dense depth sensors, such as Kinect, \cite{Izadi:2011:KRR:2047196.2047270} and \cite{6130321} proposed real-time solutions to reconstruct a 3D map which can be further down-sampled to generate the final point cloud with a reasonable resolution.

\begin{figure}[t!]
\centering
\includegraphics[width=0.7\linewidth]{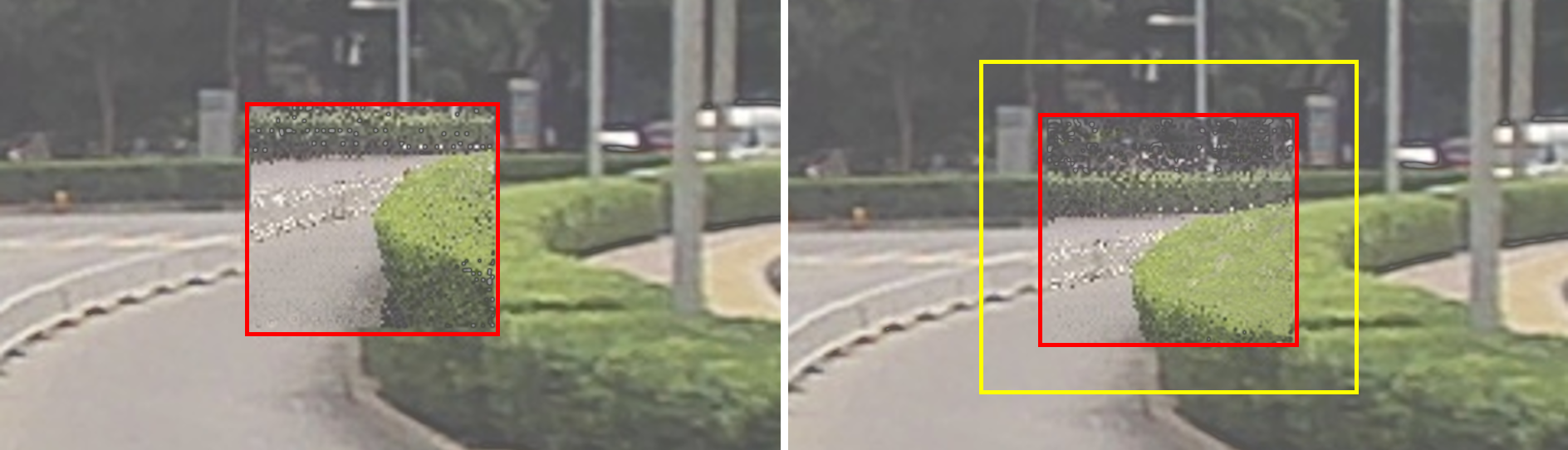}
   \caption{The point cloud is projected into the target region (red box) with colors. The left image shows projection by calibration result. Obvious misalignment can be seen at boundaries. The right image shows projection by optimized rotation \textbf{R}, where points match much better with surrounding pixels. The region between yellow and red boxes is where we compare colors of projected 3D points and image pixels to optimize camera rotation matrix \textbf{R}.
}
\label{fig:misalignment}
\end{figure}

\textbf{Camera Pose Refinement}. The relative poses between depth sensor and image sensor can be calibrated in advance, but there are still some misalignments between the point cloud and image pixels, as shown in Fig.~\ref{fig:misalignment}. Vibrations, inaccurate synchronization, and accumulative errors from point cloud stitching cause pose offset between the image sensor and depth sensor.

In order to produce seamless inpainting results, such offset should be compensated even if it's minor in most times. From the initial extrinsic calibration between the image sensor and depth sensor, we optimize their relative rotation R and translation T by minimizing the photometric projection error. The error is defined as:
\begin{equation}
    E = \sum_{p\in\Omega}| c(p) - c(q) |^2,
	\label{equ:termCor}
\end{equation} 
where $p$ is a pixel projection from 3D map. $\Omega$ is an area surrounding the target inpainting region, which is illustrated in Fig.~\ref{fig:misalignment} as the region between red and yellow boxes. $q$ is original pixel in the image overlaid by $p$. The function $c(\cdot)$ returns the value of a pixel. 

Note that the colors and locations of a pixel are discrete values, making the error function $E$ not continuous on $\mathbf{R}$ and $\mathbf{T}$. We can't solve the following equation directly using the standard solvers, such as Levenberg-Marquardt algorithm or Gauss-Newton algorithm. Instead, we search discrete spaces of $\mathbf{R}$ and $\mathbf{T}$ to minimize $E$. However, $\mathbf{R}$ and $\mathbf{T}$ have 6 degrees of freedom (DoF) in total, making the searching space extremely large. We choose to fix $\mathbf{T}$ and only optimize $\mathbf{R}$ because $\mathbf{R}$ is dominant at determining projection location when the majority of the 3D map are distant points. Moreover, in most cases, we only need to move projection pixels slightly in vertical and horizontal directions in the image space, which are determined by pitch and yaw angles of the camera. We finally reduce our search space to 2 DoF, which significantly speed up the optimization process.

\textbf{Depth Map}. Once the camera pose is refined, we project the 3D map onto each image frame to generate the corresponding depth map. Note that some point clouds are captured far from the current image, which can be occluded and de-occluded during the projection process. Hence, we employ z-buffer to get the nearest depth.

\begin{figure}[t!]
\centering
\includegraphics[width=0.35\linewidth]{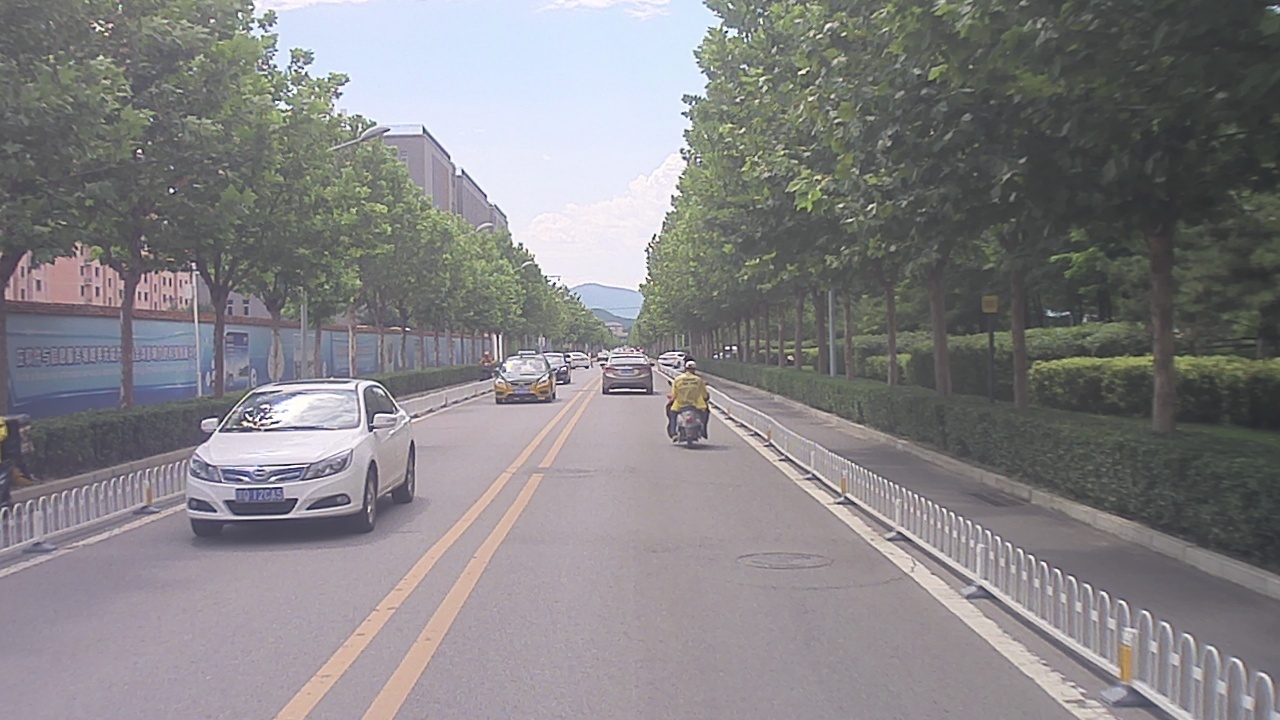}
\includegraphics[width=0.35\linewidth]{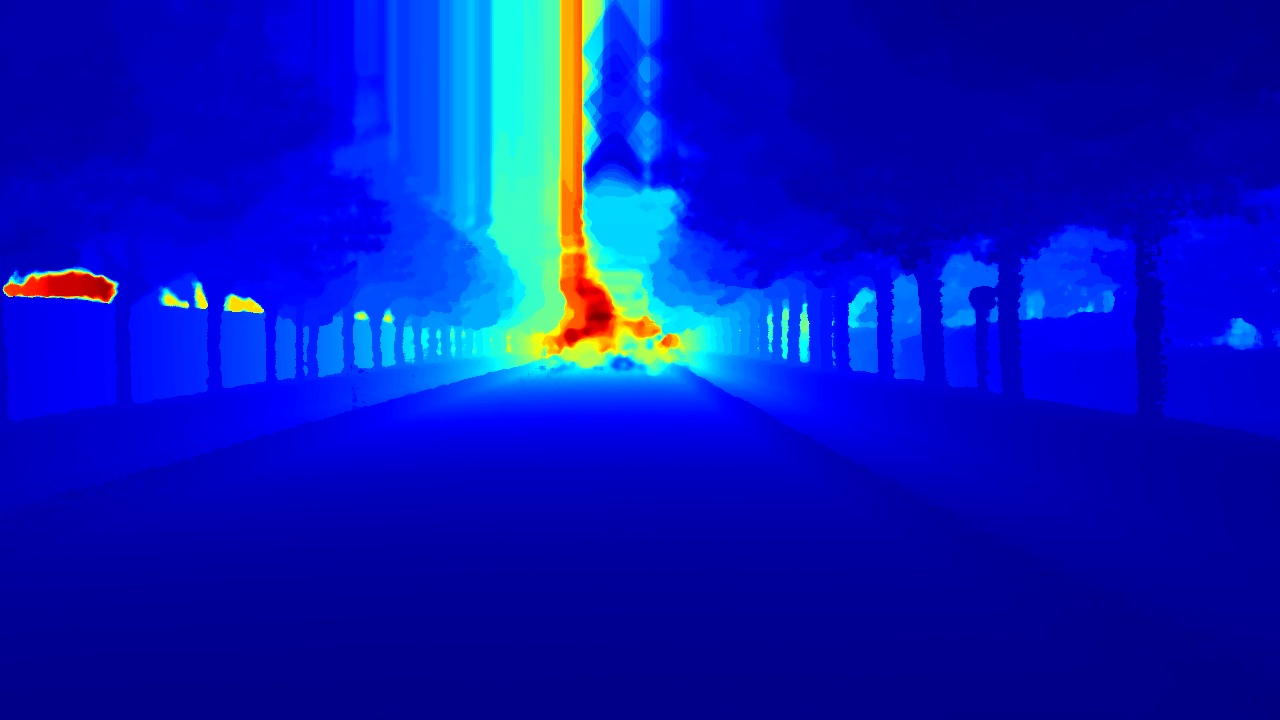}
   \caption{A color image and its corresponding dense depth map. Note that the depth is only rendered for background points and all moving objects have been removed.}
\label{fig:depthmap}
\end{figure}

To get a fully dense depth map, we could definitely borrow some of the fancy algorithms (e.g. \cite{cheng2018depth}) that learn to produce dense depth maps from sparse ones. However, we find that the simple linear interpolation is good enough to generate the dense 3D map in our cases. We further apply a $5\times5$ median filter to remove some individual noise points. The final depth map is shown in Fig.~\ref{fig:depthmap}.

\subsection{Candidate Color Sampling Criteria}

\begin{figure}[t!]
\centering
\includegraphics[width=1.0\linewidth]{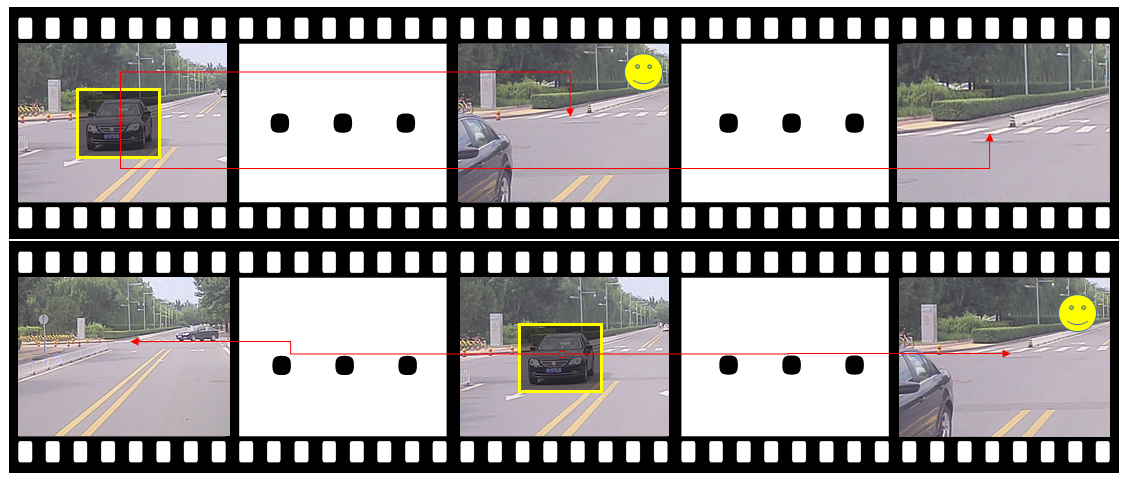}
   \caption{Color candidate selection criteria. Top row: a pixel finds its candidate colors in 2 later frames where road texture appears clear in both images. In this case, we choose the frame that is temporally close to the current frame, in order to minimize the impact of perspective change and potential occlusion or de-occlusion. Bottom row: a pixel finds its candidate colors in one previous frame and one later frame. In this case, we prefer the later frame over the previous one, since road texture is lost in the previous frame.}
\label{fig:colorsel}
\end{figure}

As every pixel is assigned a depth value, it is possible to map a pixel from one image to other images. There are multiple choices of colors to fill in the pixels of the target inpainting region, a guideline should be followed to find the best candidate color. We have 2 principles to choose the best color candidate: 1) always choose from the frame that is closer to the current frame temporally and 2) always choose from the frame where the 3D background is closer to the camera. Please refer to Fig.~\ref{fig:colorsel} for an example of our candidate selection criteria. The first requirement ensures our inpainting approach suffers less from perspective distortion and occlusion. And the second requirement is because image records more texture details when it's closer to objects, so that we can retain more details in the inpainting regions.

Under this guideline and the fact that sensors only move forwards during capture, our algorithm works by first searching forwards temporally to the end of video and then backwards until beginning. The first valid pixel is chosen as the candidate. And the valid pixel means its location doesn't fall into the target inpainting regions. 

\subsection{Regularization with Belief Propagation}

At this point, every pixel gets color value individually. If the camera pose and depth value are 100\% correct, we can generate perfect inpainting results with smooth boundaries and neighbors. However, it's not the case in the real world, especially, the depth map always carries errors. Therefore, we have to explicitly enforce some smoothness constraints. 

\begin{figure}[t!]
\centering
\includegraphics[width=1.0\linewidth]{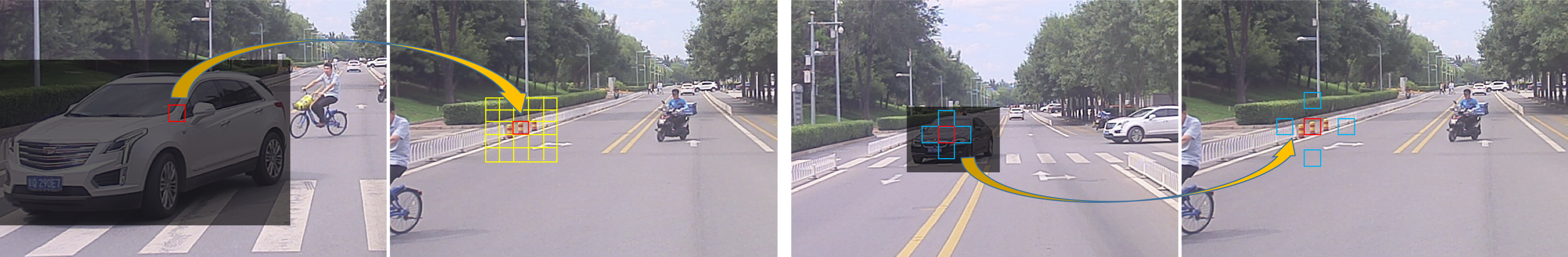}
   \caption{Left image pair: potential color choices for a target pixel. a pixel within inpainting region find its candidate pixel (red box) from a source image. A small window of pixels around this candidate (yellow boxes) are all potential colors to fill the target pixel. Right image pair: the 4 neighbors of a target pixel are not necessarily neighbors in the source image due to perspective change. In order to get neighbor colors, we need to warp neighbor pixels into the source image by their depth value. }
\label{fig:neighbors}
\end{figure}

We formulate the color selection as a discrete global optimization problem and solve it using belief propagation (BP). Before explaining the formulation, we first define the color space and neighbors of a target pixel. As shown in left image pair in Fig.~\ref{fig:neighbors}, a target pixel (left red box) finds its candidate pixel (right red box) from a source image, due to depth inaccuracy, the true color might not lie exactly on the candidate pixel, but a small window around. So we collect all pixel colors from this small n by n window to form the color space for the target pixel. The right image pair in Fig.~\ref{fig:neighbors} illustrates how to find out the expected colors of neighbors. Because of perspective changes, the 4 neighbors of a target pixel are not necessarily neighbors in the source image. Hence, we warp neighbor pixels into the source image by their depth value to sample the expected colors.

Let $P$ be the set of pixels in the target inpainting region and L be a set of labels. The labels correspond to the indices of potential colors in the color space. A labeling function l assigns a $l_p \in L$ to each pixel $p \in P$. We assume that the labels should vary smoothly almost everywhere but may change dramatically at some places such as object boundaries. The quality of labeling is given by an energy function as 
\begin{equation}
E = \sum_{(p, q) \in N}V(l_p, l_q) + \sum_{p \in P}D_p(l_p),
\end{equation} 
where $N$ are the number of edges in the four-connected image grid graph. $V(l_p, l_q)$ is the cost of assigning labels $l_p$ and $l_q$ to two neighboring pixels, and is normally referred to as the discontinuity cost. $D_p(l_p)$ is the cost of assigning label $l_p$ to pixel $p$,  which is referred to as the data cost. Determining a labeling with minimum energy corresponds to the Maximum A Posteriori (MAP) estimation problem.

We incorporate boundary smoothness constraint into the data cost as following:
\begin{equation}
D_p(l_p) = 
\begin{cases}
|C_{pl}(l_p) - I(q)|, & \text{if p is left boundary pixel} \\
|C_{pr}(l_p) - I(q)|, & \text{if p is right boundary pixel} \\
|C_{pt}(l_p) - I(q)|, & \text{if p is top boundary pixel} \\
|C_{pb}(l_p) - I(q)|, & \text{if p is bottom boundary pixel} \\
\alpha, & \text{otherwise}
\end{cases} ,
\end{equation}
where $C_{pl}, C_{pr}, C_{pt}, C_{pb}$ return expected colors of pixel $p$'s left, right, top and bottom neighbors respectively. $q$ is the neighbor pixel of $p$ outside of the inpainting region in the target image, so it has known color, which is returned by the function $I(q)$. Here we take the difference of true neighbor color and expected neighbor color as a measure of labeling quality. For those pixels not on the inpainting boundary, we give equal opportunities to all the labels by assigning a constant value of $\alpha$. The discontinuity cost is defined as
\begin{equation}
V(l_p, l_q) = 
\begin{cases}
|C_{pl}(l_p)-C_{q}(l_q)|+|C_p(l_p)-C_{qr}(l_q)| & \text{if L } \\
|C_{pr}(l_p)-C_{q}(l_q)|+|C_p(l_p)-C_{ql}(l_q)| & \text{if R } \\
|C_{pt}(l_p)-C_{q}(l_q)|+|C_p(l_p)-C_{qb}(l_q)| & \text{if T } \\
|C_{pb}(l_p)-C_{q}(l_q)|+|C_p(l_p)-C_{qt}(l_q)| & \text{if B}
\end{cases} .
\end{equation}
Here, $C_p(\cdot)$ and $C_q(\cdot)$ fetch colors for $p$ and $q$ at label $l_p$ and $l_q$. $L, R, T, B$ stand for $q$ is on left, on right, on top and on bottom respectively. For a pair of 2 neighboring pixels $p$ and $q$, we compute differences between $p$'s color and $q$'s expected color of $p$ and vice versa.

% The problem is solved by belief propagation, so that we have enforce both boundary smoothness and neighbor smoothness. In later section, we will show the effectiveness of applying this global optimization.

% \subsection{Optional Post Processing}
% Our algorithm has an implicit assumption that the inpainting regions must be visible in some other frames. Otherwise, some pixels will remain blank, as can be seen from figure~\ref{fig:comp1}. In this case, We could use those GAN based approaches~\cite{IizukaSIGGRAPH2017, pathakCVPR16context, yu2018generative, Xu_2019_CVPR} to fill the remain pixels. This makes the task much easier for those approaches because remaining holes are much smaller. The results are much better than applying them directly onto the original holes. We will demonstrate this in next section.

\begin{figure}[t!]
\centering
\includegraphics[width=1\linewidth]{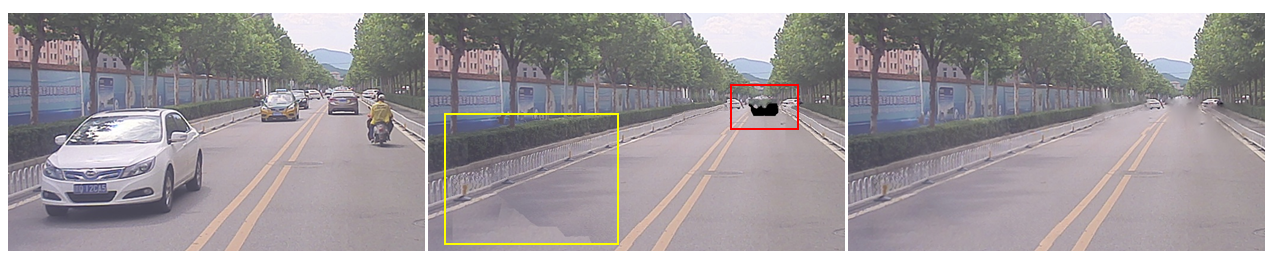}
   \caption{Left image: input image. Middle image: inpainting result. Note the color discontinuity in yellow box and blank pixels in red box. Right image: result after color harmonization. }
\label{fig:colorharmo}
\end{figure}

\subsection{Color Harmonization}
Pixels from different frames may have different colors due to changing camera exposure time and white balance, causing color discontinuities (Fig.~\ref{fig:colorharmo}). We borrow Poisson image editing~\cite{Perez:2003:PIE:1201775.882269} to solve these problems. Poisson image editing is originally proposed to clone an image patch from source image into a destination image with seamless boundary and original texture. It achieve this by solving the following minimization problem 
\begin{equation}
    \min_f{\int\int_\Omega|\Delta f-v|}  \text{   with   } f|_{\partial\Omega}=f^*|_{\partial\Omega}.
	\label{equ:poisson}
\end{equation}
Here all the notations are inherited from~\cite{Perez:2003:PIE:1201775.882269}. $\Omega$ is the inpainting region with boundary $\partial\Omega$. $f^*$ is color function of destination image and $f$ is color function of the target inpainting region within destination image. $\Delta.=[\partial./\partial x, \partial./\partial y]$ is gradient operator. $v$ is the desired color gradient defined over $\Omega$.

In our case, $v$ is computed using the output from the belief propagation step, with one exception. If two neighboring pixels within $\Omega$ are from different frames, we set their gradient to 0. This guarantee color consistency within the inpainting regions. The effectiveness of this solution is demonstrated in Fig.~\ref{fig:colorharmo}. Note that the blank-pixel region is also filled up. Since blank pixels have 0 gradient values, solving the Poisson equation on this part is equivalent to smooth color interpolation.

\subsection{Video Fusion} 
Our algorithm has an implicit assumption that the inpainting regions must be visible in some other frames. Otherwise, some pixels will remain blank, as can be seen from Fig.~\ref{fig:colorharmo}. Learning-based methods can hallucinate inpainting colors from their training data. In contrast, our approach can't inpaint occluded areas if they are never visible in the video, leaving blank pixels.   

For small areas of blank pixels, a smooth interpolation is sufficient to fill the hole. However, in some cases, a vehicle in front could block a wide field of view for the entire video duration, leaving big blank holes. A simple interpolation will not be capable of handling this problem. A better way to address this issue would be capturing another video of the same scene, where the occluded parts become visible. Fortunately, it is straightforward to register newly captured frames into an existing 3D map using LOAM~\cite{Zhang-2014-7903}. Once new frames are registered and merged into the existing 3D map, inpainting is performed exactly the same way. Some of our results of video fusion can be found in the next section as well as in supplemental materials.

\subsection{Temporal Smoothing}
Finally, we compute both forward and backward optical flows for all the result frames. For every pixel in the target inpainting areas, we trace it into neighboring frames using the optical flow and replace its original color with average of colors sampled from neighbor frames. 

%% file: 4-exp.tex
\section{Experiments and Results}

\begin{figure}[b!]
%\centering
\begin{tabular}{@{}c@{}}
    \centering
    \raisebox{-.5\height} {
    \includegraphics[width=0.167\linewidth]{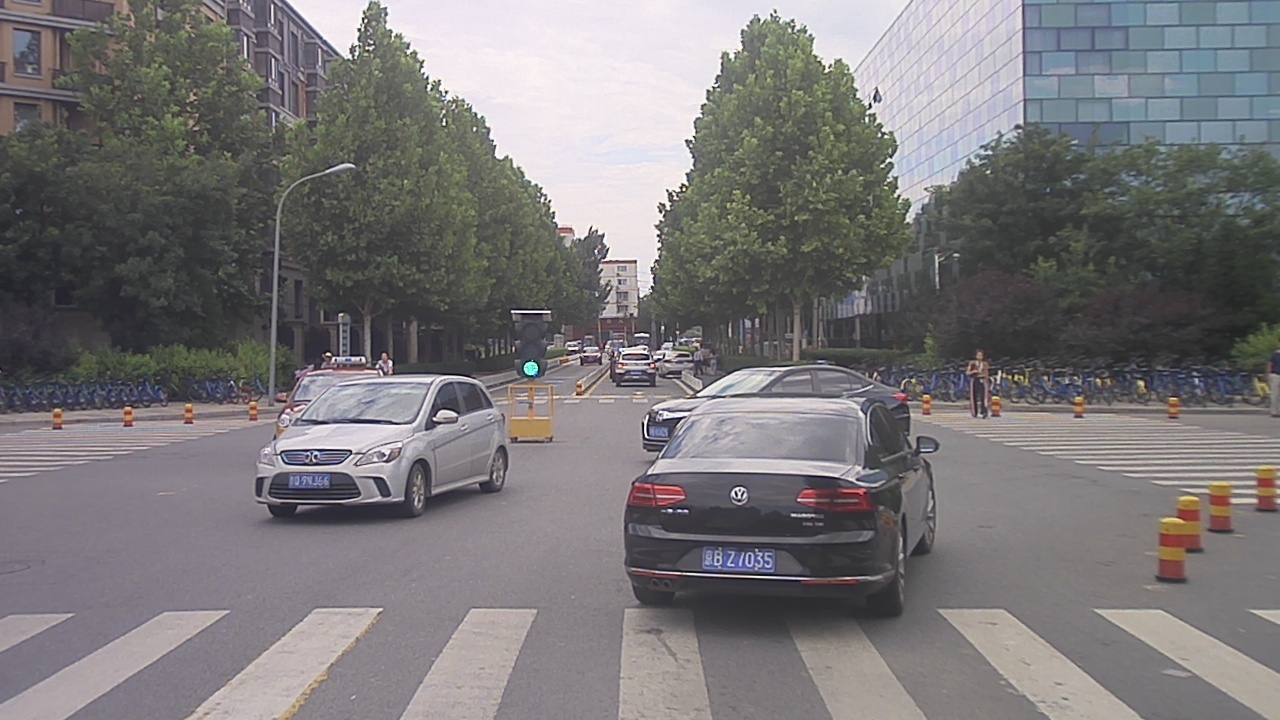}\hspace{-0.06 cm}
    \includegraphics[width=0.167\linewidth]{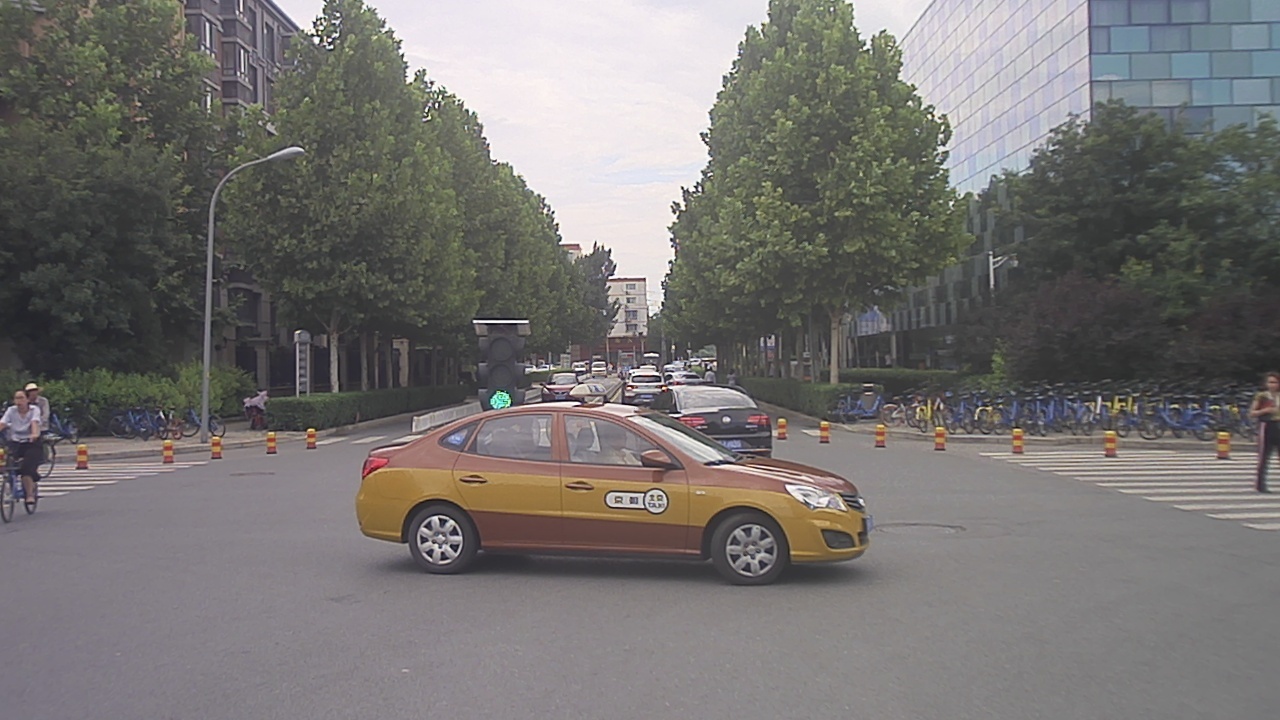}\hspace{-0.06 cm}
    \includegraphics[width=0.167\linewidth]{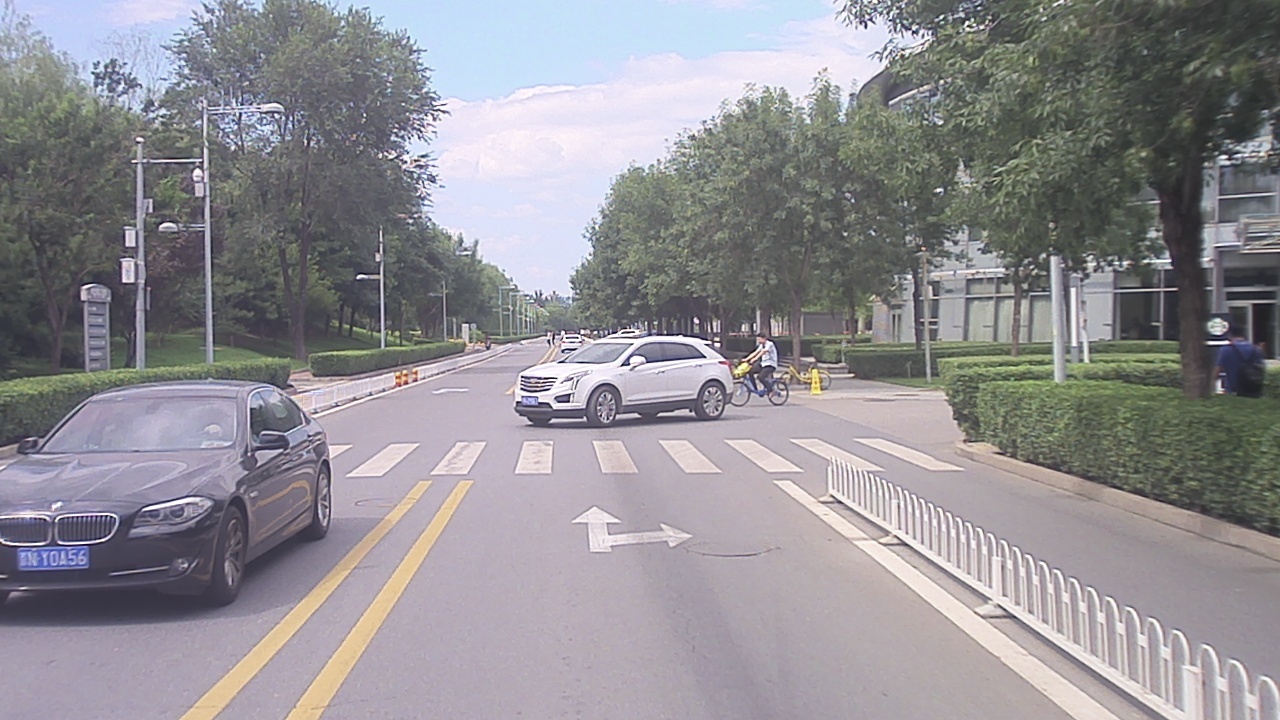}\hspace{-0.06 cm}
    \includegraphics[width=0.167\linewidth]{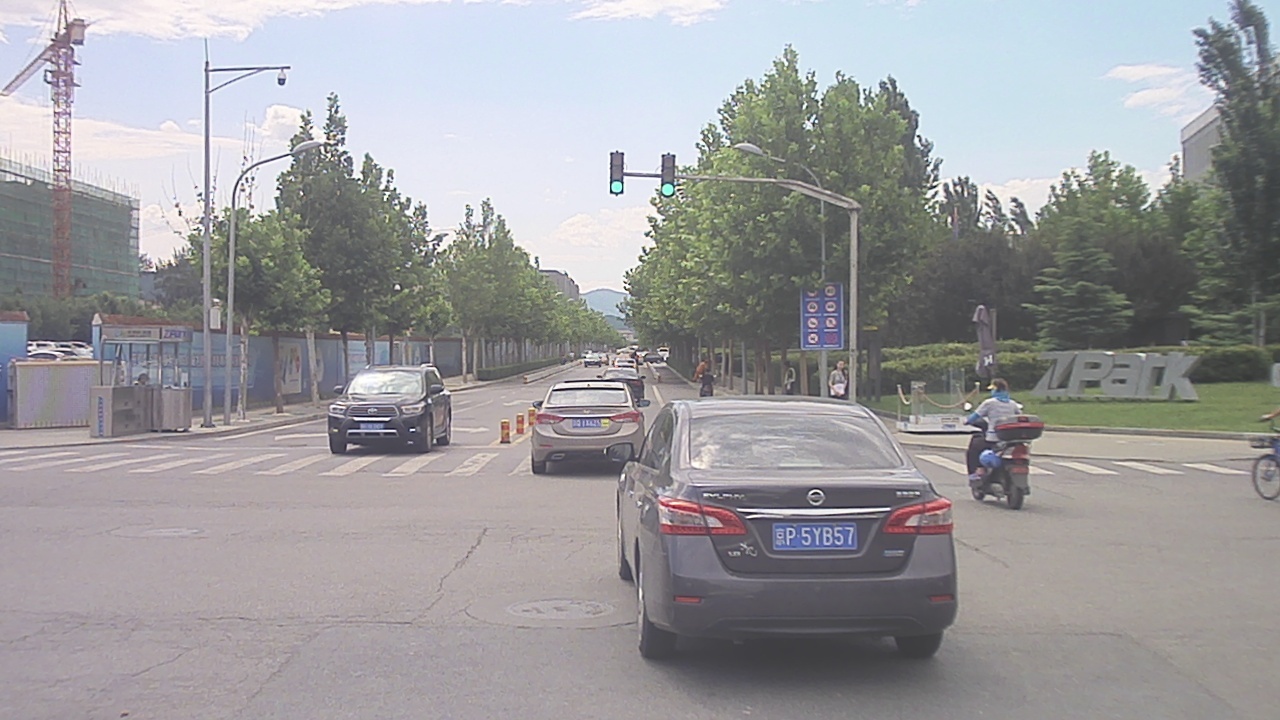}\hspace{-0.06 cm}
    \includegraphics[width=0.167\linewidth]{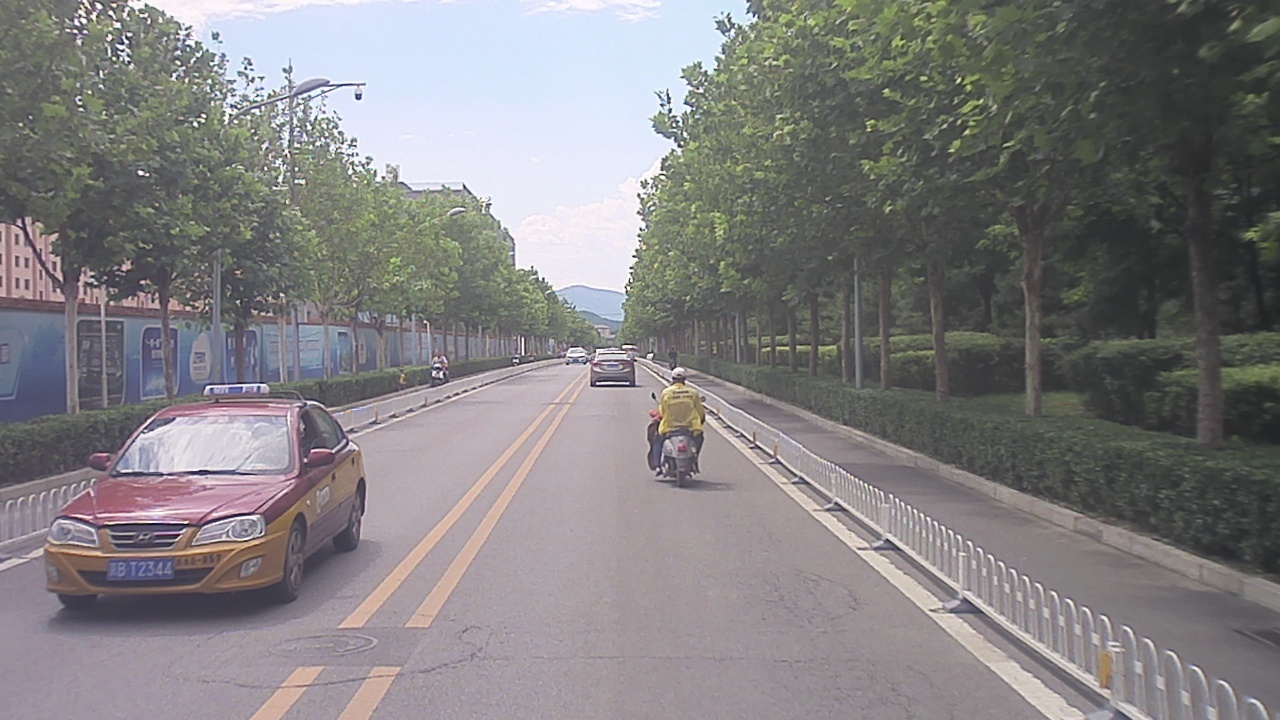}
    } Inputs
\end{tabular}

\begin{tabular}{@{}c@{}}
    \centering
    \raisebox{-.5\height} {
    \includegraphics[width=0.167\linewidth]{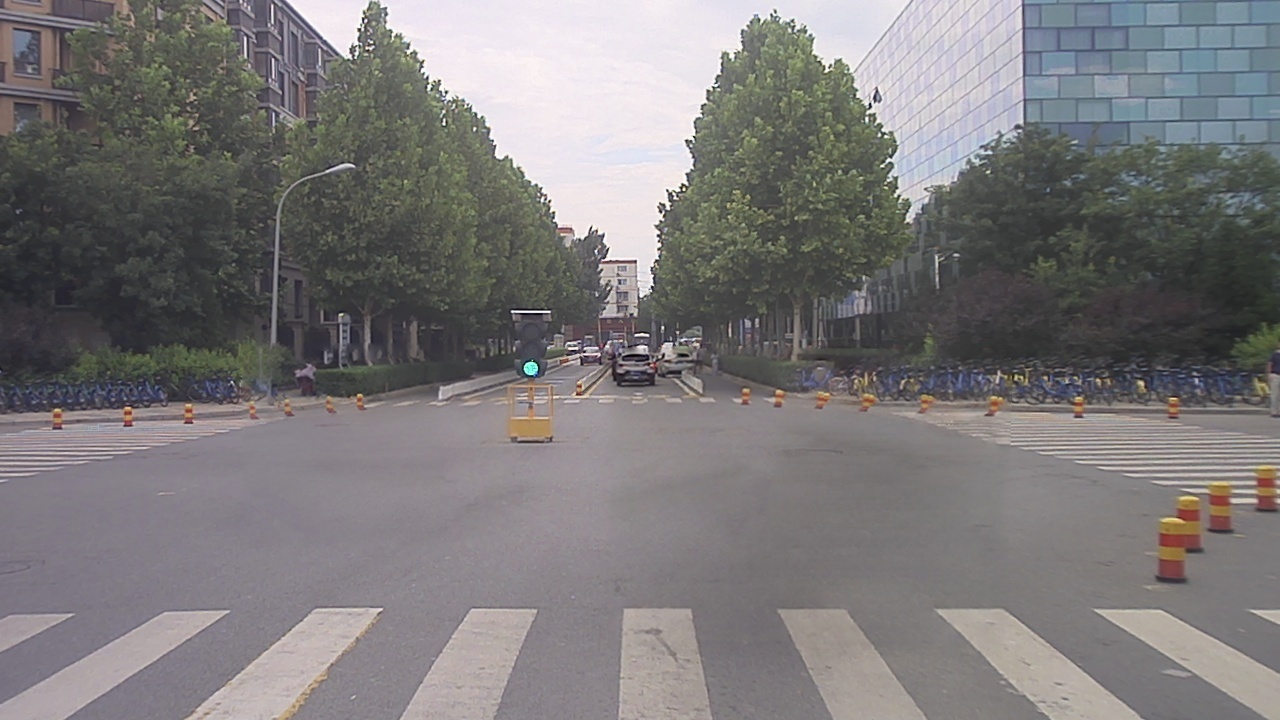}\hspace{-0.06 cm}
    \includegraphics[width=0.167\linewidth]{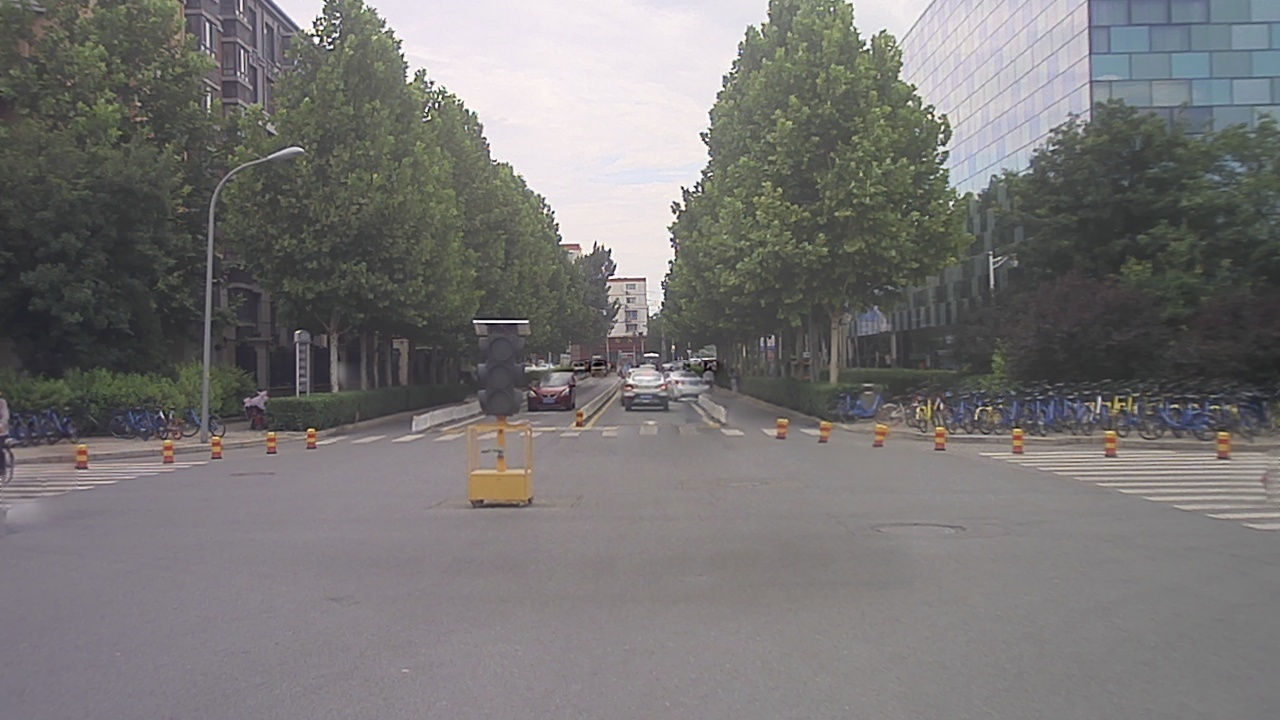}\hspace{-0.06 cm}
    \includegraphics[width=0.167\linewidth]{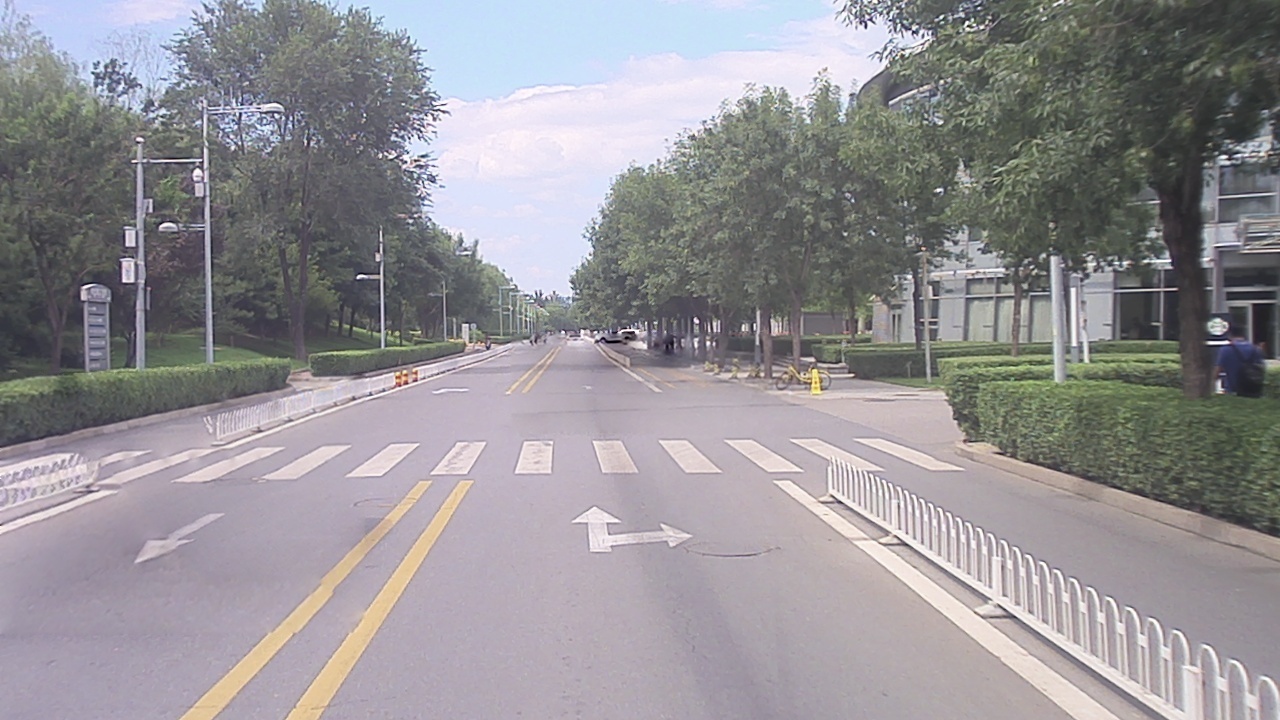}\hspace{-0.06 cm}
    \includegraphics[width=0.167\linewidth]{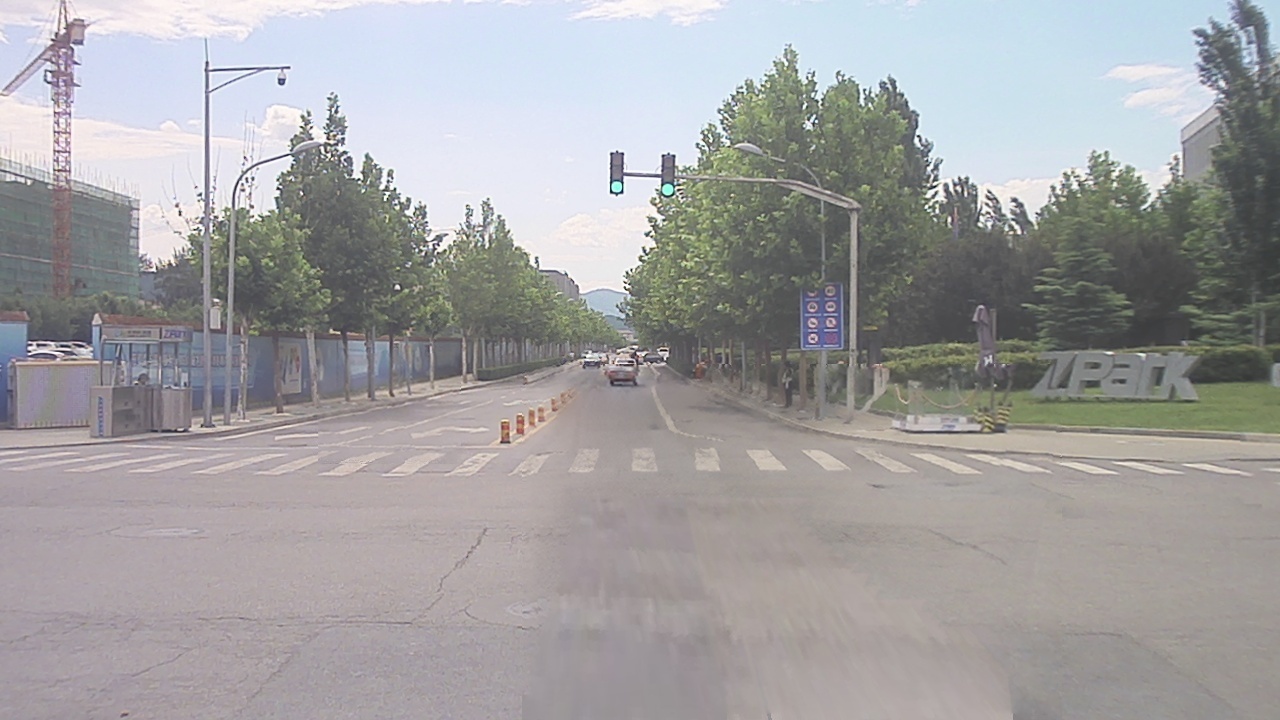}\hspace{-0.06 cm}
    \includegraphics[width=0.167\linewidth]{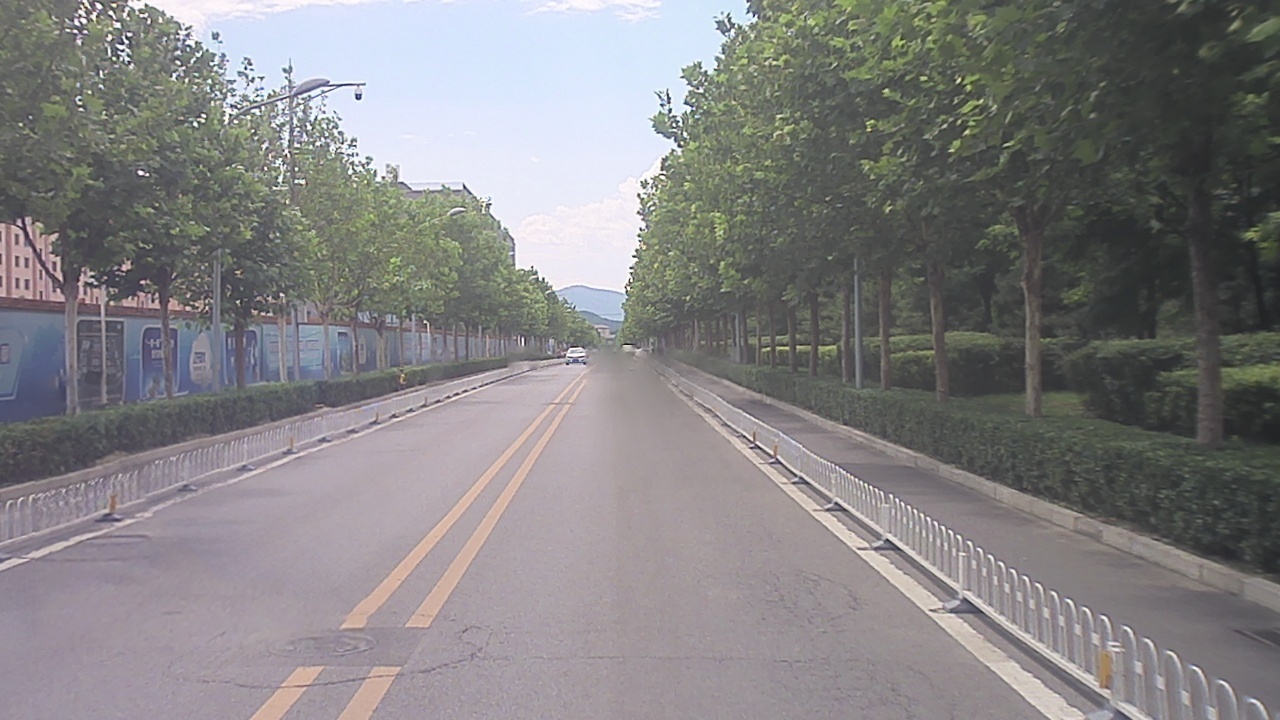}
    }  Ours 
\end{tabular}

\begin{tabular}{@{}c@{}}
    \centering
    \raisebox{-.5\height} {
    \includegraphics[width=0.167\linewidth]{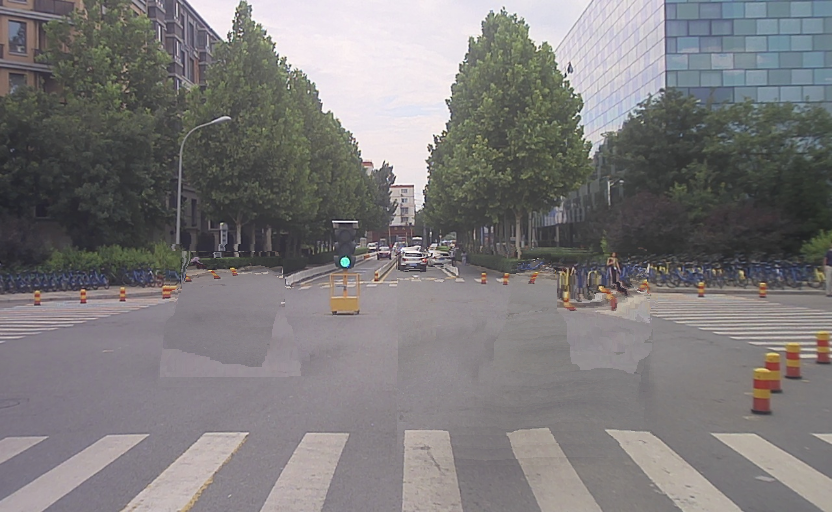}\hspace{-0.06 cm}
    \includegraphics[width=0.167\linewidth]{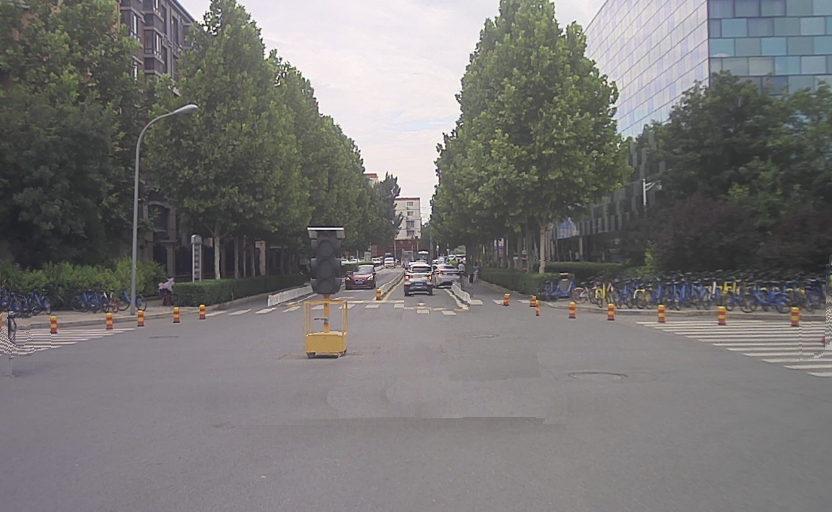}\hspace{-0.06 cm}
    \includegraphics[width=0.167\linewidth]{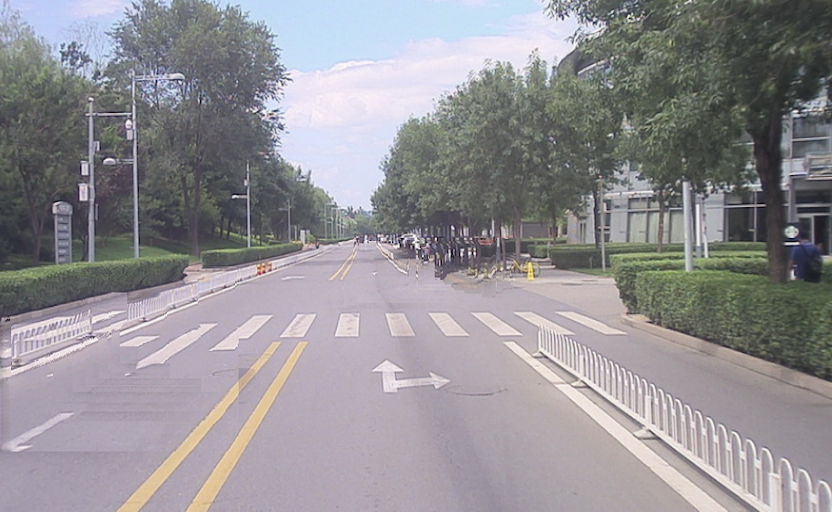}\hspace{-0.06 cm}
    \includegraphics[width=0.167\linewidth]{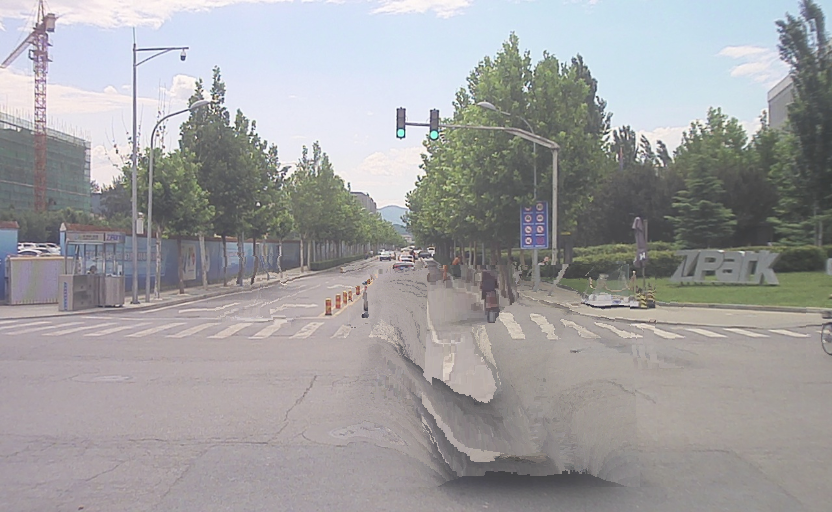}\hspace{-0.06 cm}
    \includegraphics[width=0.167\linewidth]{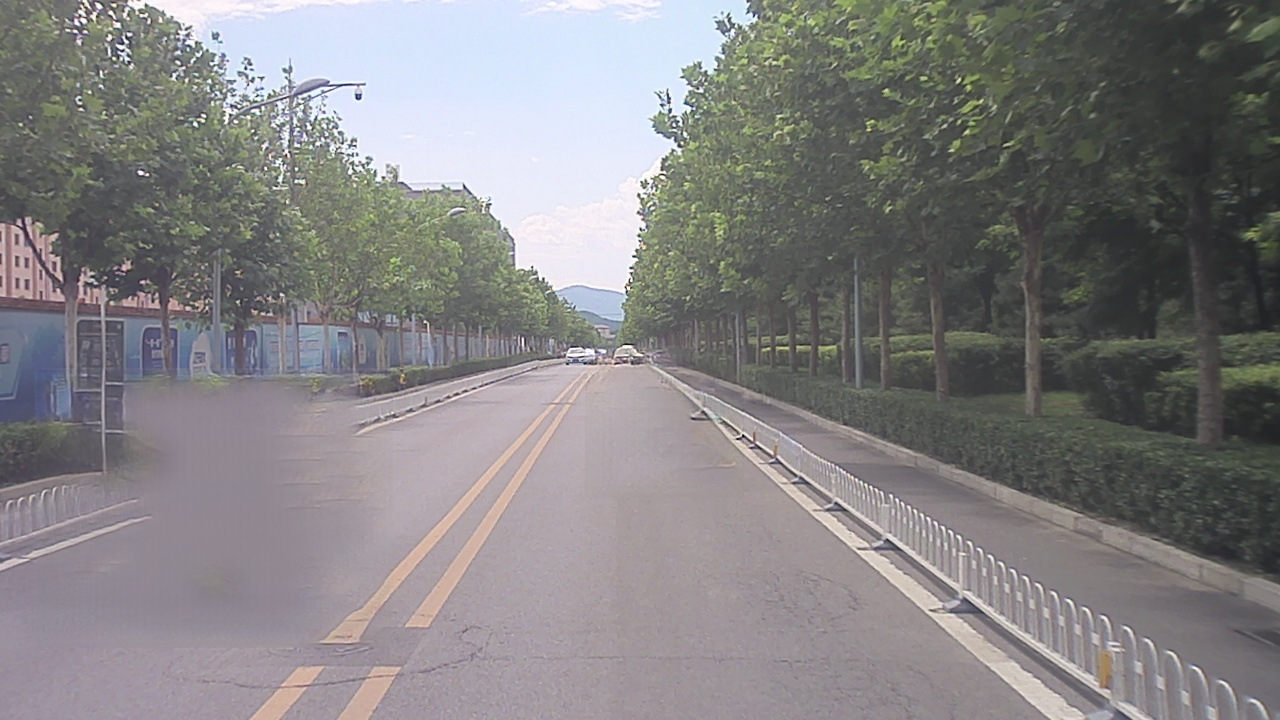}
    } Yu~\cite{yu2018generative}
\end{tabular}

\begin{tabular}{@{}c@{}}
    \centering
    \raisebox{-.5\height} {
    \includegraphics[width=0.167\linewidth]{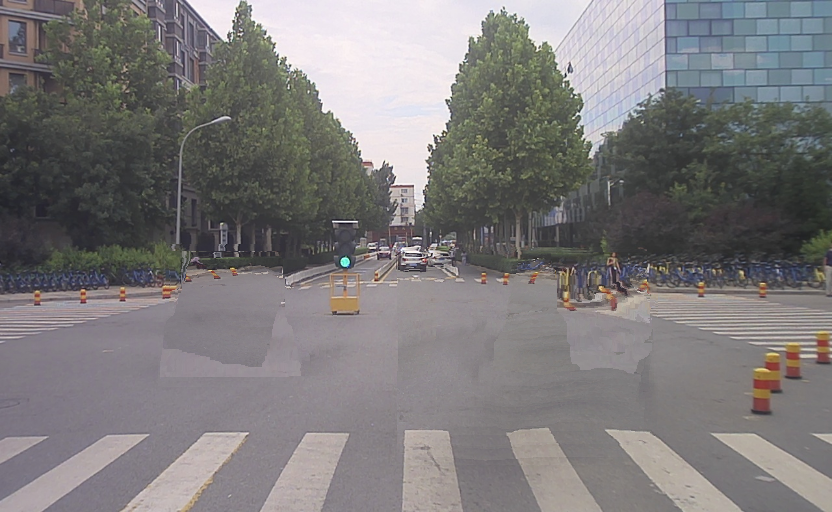}\hspace{-0.06 cm}
    \includegraphics[width=0.167\linewidth]{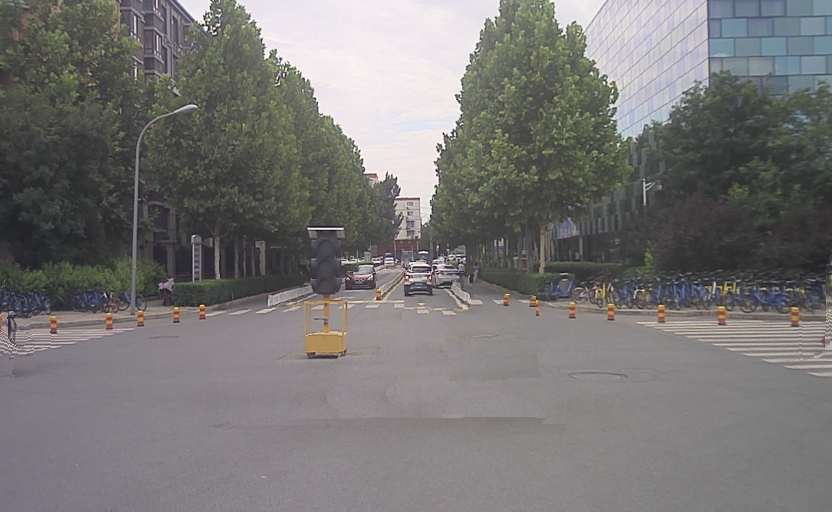}\hspace{-0.06 cm}
    \includegraphics[width=0.167\linewidth]{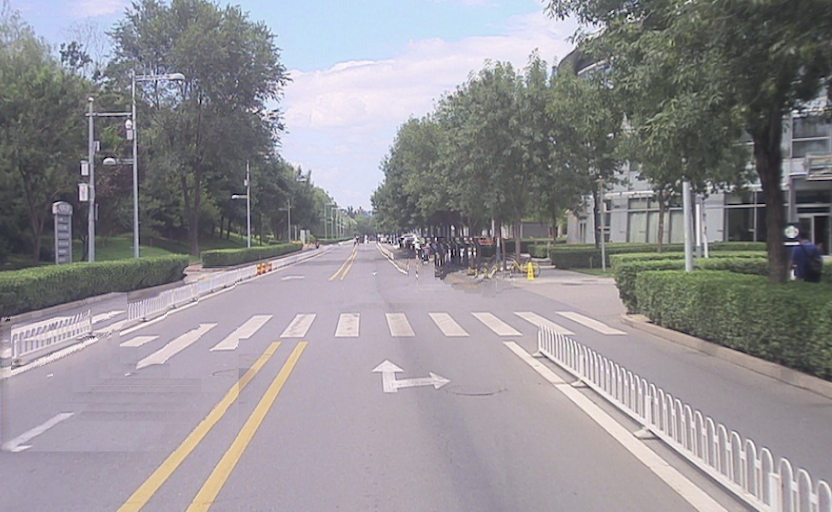}\hspace{-0.06 cm}
    \includegraphics[width=0.167\linewidth]{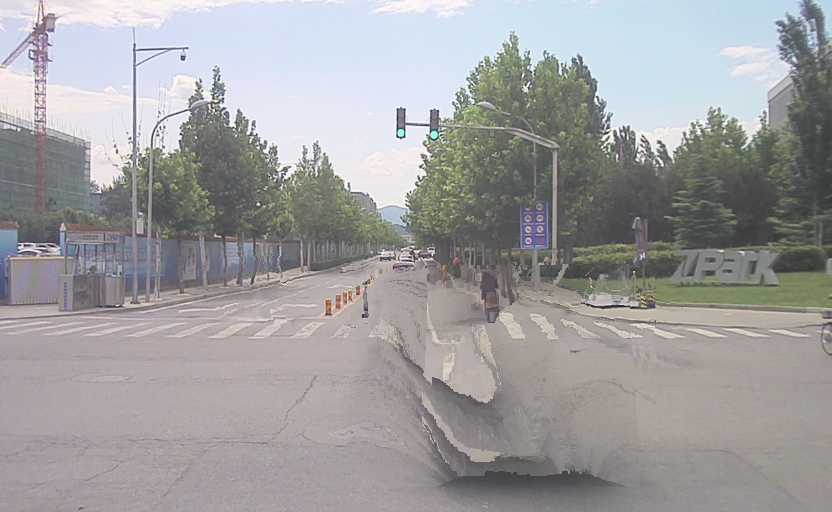}\hspace{-0.06 cm}
    \includegraphics[width=0.167\linewidth]{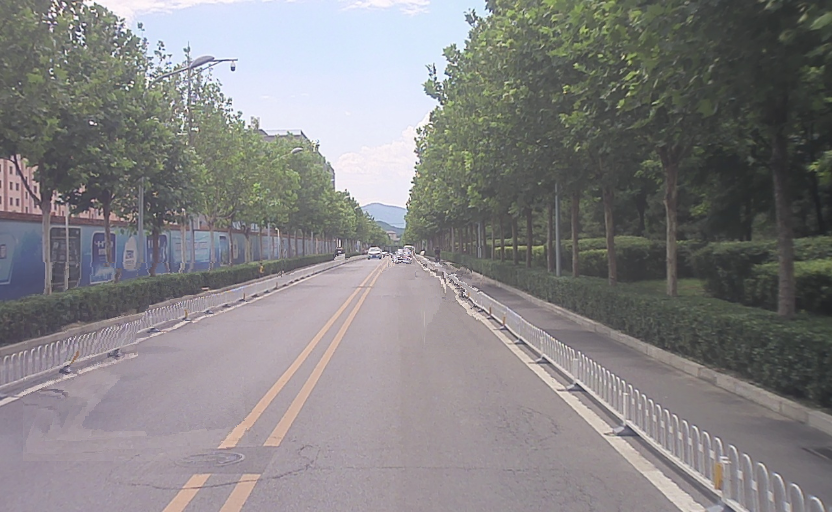}
    } Xu~\cite{Xu_2019_CVPR}
\end{tabular}

\begin{tabular}{@{}c@{}}
    \centering
    \raisebox{-.5\height} {
    \includegraphics[width=0.167\linewidth]{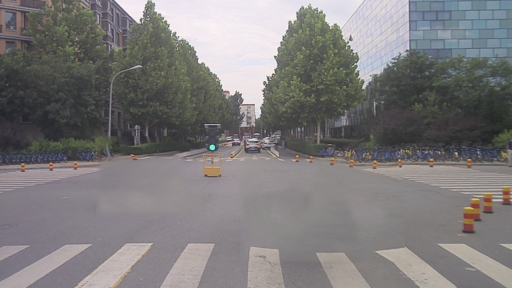}\hspace{-0.06 cm}
    \includegraphics[width=0.167\linewidth]{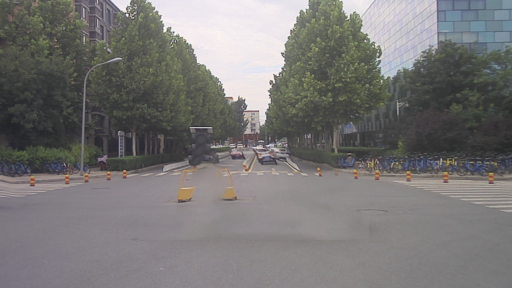}\hspace{-0.06 cm}
    \includegraphics[width=0.167\linewidth]{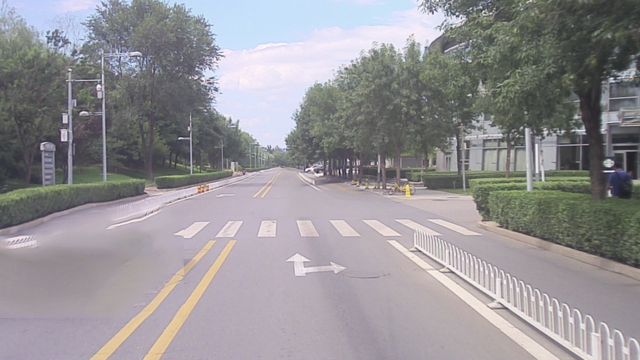}\hspace{-0.06 cm}
    \includegraphics[width=0.167\linewidth]{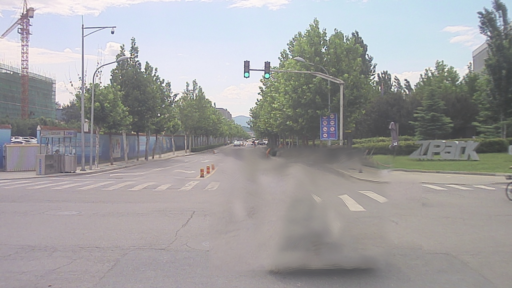}\hspace{-0.06 cm}
    \includegraphics[width=0.167\linewidth]{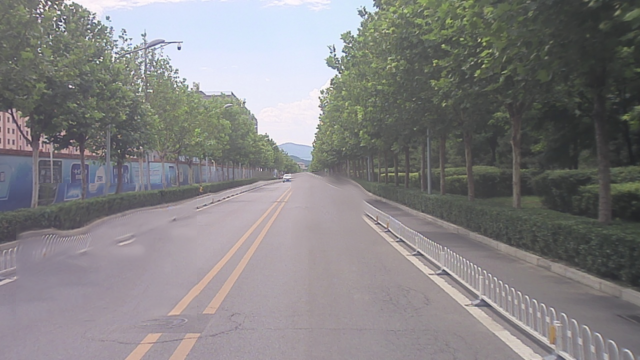}
    } Huang~\cite{Huang-SigAsia-2016}
\end{tabular}

\caption{5 frames from different video clips are demonstrated to compare our results with others.}
\label{fig:comp2}
\end{figure}

To our best knowledge, all the public datasets (including  DAVIS Dataset~\cite{perazzi2016benchmark}) for video inpainting don't come with depth, which is a must for our algorithm. Autonomous driving dataset ApolloScape~\cite{ma2019trafficpredict} indeed have both camera images and point clouds, but it's not adopted by research community to evaluate video inpainting. Plus, its dataset was captured by a professional mapping Lidar RIEGL, which is not a typical setup for an autonomous driving car. Thus, we captured our own dataset and compare to previous works on our dataset.

\subsection{Inpainting Dataset}

We use an autonomous driving car to collect large-scale datasets in urban streets. The data is generated from a variety of sensors, including Hesai Pandora all in one sensor (40-beam LiDAR, 4 mono cameras covering 360 degrees, 1 forward-facing color camera), and a localization system working at 10 HZ. The LiDAR is synchronized with embedded frontal facing wide-angle color camera. We recorded a total of 5 hours length of RGB videos includes 100K synchronized $1280\times720$ images and point cloud. The dataset includes many challenging scenes e.g. background is occluded by large bus, shuttle or truck in the intersection and the front car is blocking the front view all the time. For those long time occlusion scenarios, the background is missing in the whole video sequence. We captured these difficult streets/intersections more than once, providing us the data for video fusion inpainting. Our new dataset will be published with the paper.

\subsection{Comparisons}

We qualitatively and quantitatively compare our results to three state-of-the-art works: two video inpainting approaches~\cite{Xu_2019_CVPR, Huang-SigAsia-2016} and one image inpainting approach~\cite{yu2018generative}. For those two deep learning-based approaches~\cite{Xu_2019_CVPR} and~\cite{yu2018generative}, we re-train their models on our dataset by randomly sampling missing regions on input frames to perform a fair comparison.

\textbf{Qualitative Comparison}. In Fig.~\ref{fig:comp2}, we compare our results with three other methods. It is clear that our method produces better results than others. Even though Huang~\cite{Huang-SigAsia-2016} got smooth inpainting results, almost all the texture details are missing in their results. As shown, Yu~\cite{yu2018generative} and Xu~\cite{Xu_2019_CVPR} sometimes fill totally messy texture in the target regions.

\begin{figure}[b!]
\centering
\includegraphics[width=0.95\linewidth]{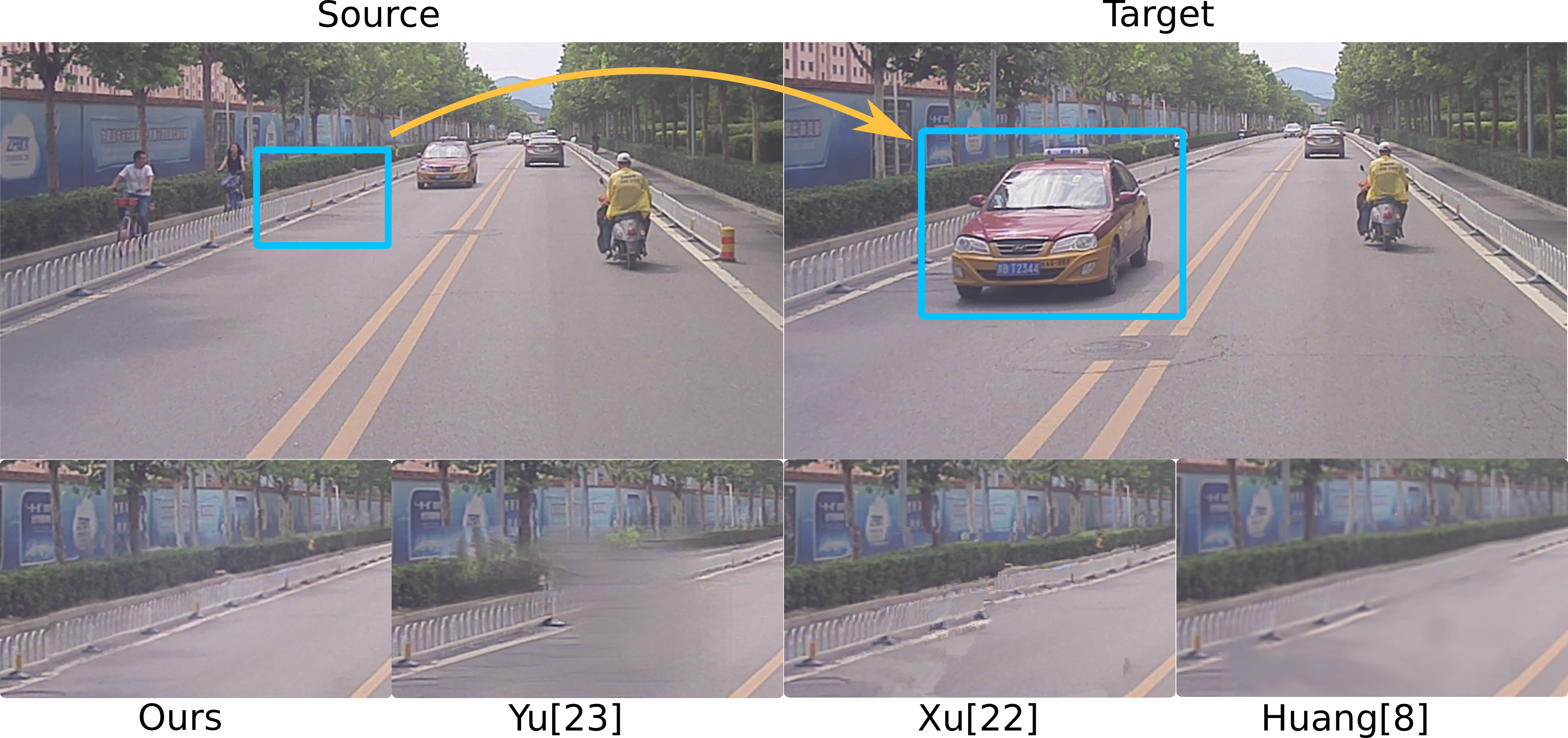}
   \caption{Top row: a patch from source image needs to be used to inpaint an occluded region in target image. Although there is significant perspective change from source to target images, our method produces geometrically and visually correct results. While other methods either fail to recover detailed texture or fail to place the texture in the right place.}
\label{fig:perspective_comp}
\end{figure}

Fig.~\ref{fig:perspective_comp} illustrates our capability to handle perspective change between source and target frames. Since our method is based on 3D geometry, perspective changes are inherently handled correctly. However, existing methods have a hard time overcoming this issue. They either fail to recover detailed texture or fail to place the texture in the right place.

%\subsection{Quantitative Comparison}

\begin{table}[]
\centering
%\resizebox{0.5\textwidth}{!}
 {%
\begin{tabular}{ l c c c c }
\hline \hline
Methods & MAE             & RMSE            & PSNR            & SSIM            \\ \hline 
Yu~\cite{yu2018generative}     & 10.961          & 16.848          & 20.821          & 0.850          \\ 
Xu~\cite{Xu_2019_CVPR}     & 7.569          & 12.932          & 19.220          & 0.594          \\  
Huang~\cite{Huang-SigAsia-2016}  & 6.924          & 11.017          & 20.022            & 0.762          \\  
Ours   & \textbf{6.135} & \textbf{9.633} & \textbf{21.631} & \textbf{0.895} \\ \hline \hline
\end{tabular}
}
\caption{Quantitative comparison with other methods, where the best results are highlighted in bold. To be clear, the values of ``MAE'' and ``RMSE'' are the lower the better while the values of ``PSNR'' and ``SSIM'' are the higher the better. }
\label{tab:comp}
\end{table}

% \begin{table}[]
% \centering
% \begin{tabular}{|c|c|c|c|c|}
% \hline
% Method & MAE             & RMSE            & PSNR            & SSIM            \\ \hline
% Xu     & 17.888          & 33.037          & 27.943          & 0.9548          \\ \hline
% Huang  & 21.438          & 32.033          & 25.271          & 0.8066          \\ \hline
% Yu     & 33.008          & 46.228          & 25.682          & 0.9432          \\ \hline
% Ours   & \textbf{16.471} & \textbf{31.905} & \textbf{28.451} & \textbf{0.9666} \\ \hline
% \end{tabular}
% \caption{Quantitative comparison to other methods.}
% \label{tab:comp}
% \end{table}

\textbf{Quantitative Comparison}. To quantitatively compare our method with other methods, we manually labeled some background areas as the target inpainting regions and use them as the ground truth. We utilize four metrics for the evaluations: Mean Absolute Error (MAE), Root Mean Squared Error (RMSE), Peak Signal to Noise Ratio (PSNR), and Structural Similarity Index (SSIM). Tab.~\ref{tab:comp} shows the evaluation results of the baseline methods and our method. Note that our method outperforms others on all four metrics. Our method reduce RMSE by 13\% compared to SOTA method.

% \begin{figure}
% % \centering
% \begin{subfigure}
% \centering
% \includegraphics[width=0.49\linewidth]{latex/images/blendingeffect/81723_input.jpg}\hspace{-0.06 cm}
% \includegraphics[width=0.49\linewidth]{latex/images/blendingeffect/80562_input.jpg}
% \end{subfigure}

% \begin{subfigure}
% \centering
% \includegraphics[width=0.49\linewidth]{latex/images/blendingeffect/81723_noblending.jpg}\hspace{-0.06 cm}
% \includegraphics[width=0.49\linewidth]{latex/images/blendingeffect/80562_noblending.jpg}
% \end{subfigure}

% \begin{subfigure}
% \centering
% \includegraphics[width=0.49\linewidth]{latex/images/blendingeffect/81723_blending.jpg}\hspace{-0.06 cm}
% \includegraphics[width=0.49\linewidth]{latex/images/blendingeffect/80562_blending.jpg}
% \end{subfigure}

% \caption{Effect of using Poisson blending. 1st row: input frames. 2nd row: results without color blending. 3rd row: results after apply Poisson color blending. }
% \label{fig:blendingeffect}
% \end{figure}

% \subsection{Ablation Study}

% \begin{figure}
% \centering
% \begin{minipage}[t]{.45\textwidth}
%   \centering
%   \includegraphics[width=0.8\linewidth]{fig/distortion1.png}
%   \caption{One frame generated by TTS audio when people pause speaking. Mouth shape is distorted.}
%   \label{fig:distorted}
% \end{minipage}%
% \hspace{1cm}%
% \begin{minipage}[t]{.45\textwidth}
%   \centering
%   \includegraphics[width=0.86\linewidth]{fig/nohand.png}
%   \caption{One frame generated by a skeleton model without hands. It is clear that hand model is necessary to render hand details in the final image.}
%   \label{fig:nohand}
% \end{minipage}
% \end{figure}

\subsection{Ablation Study}

\begin{figure*}
\centering

    \begin{minipage}{\textwidth}
    \centering
    \includegraphics[width=0.22\linewidth]{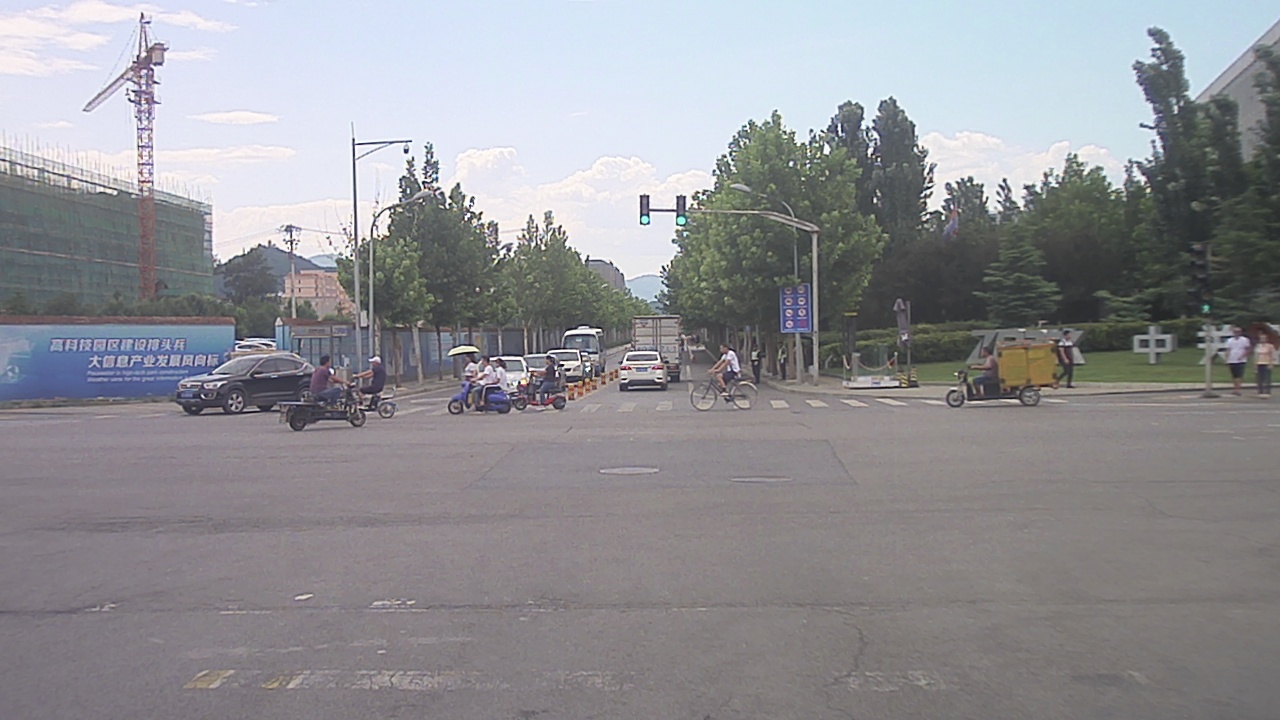}\hspace{-0.06 cm}
    \includegraphics[width=0.22\linewidth]{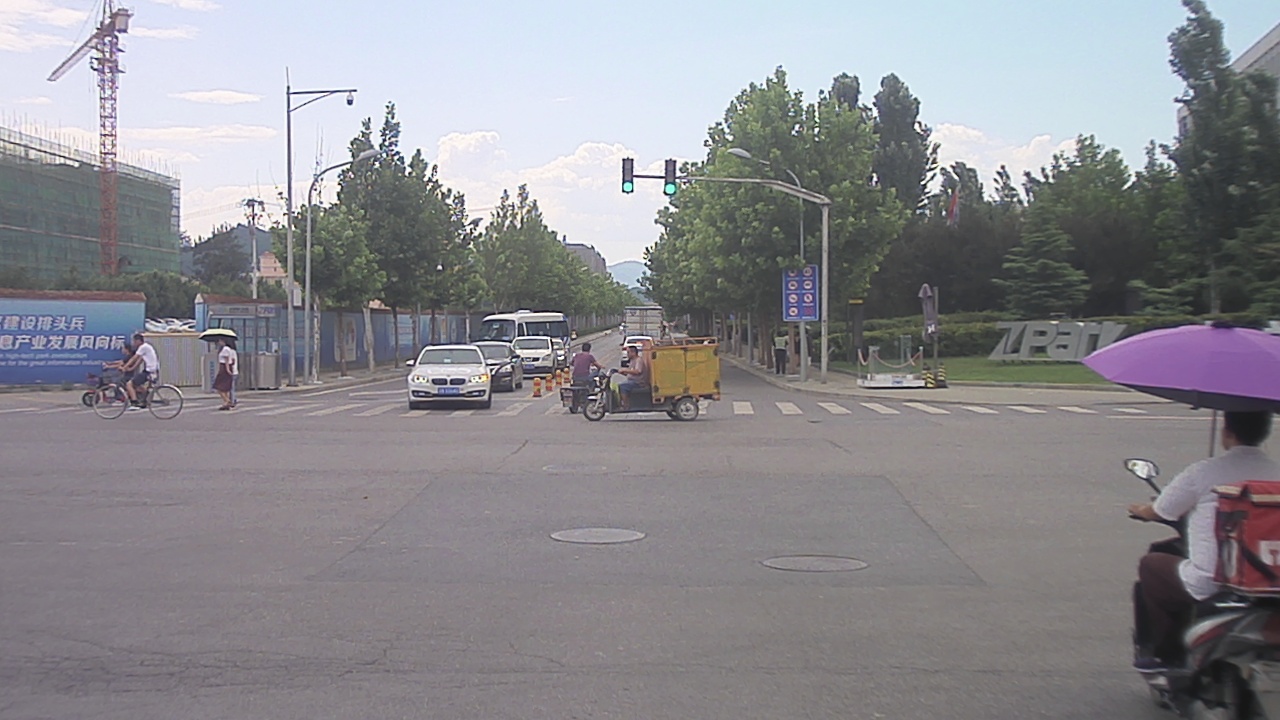}\hspace{-0.06 cm}
    \includegraphics[width=0.22\linewidth]{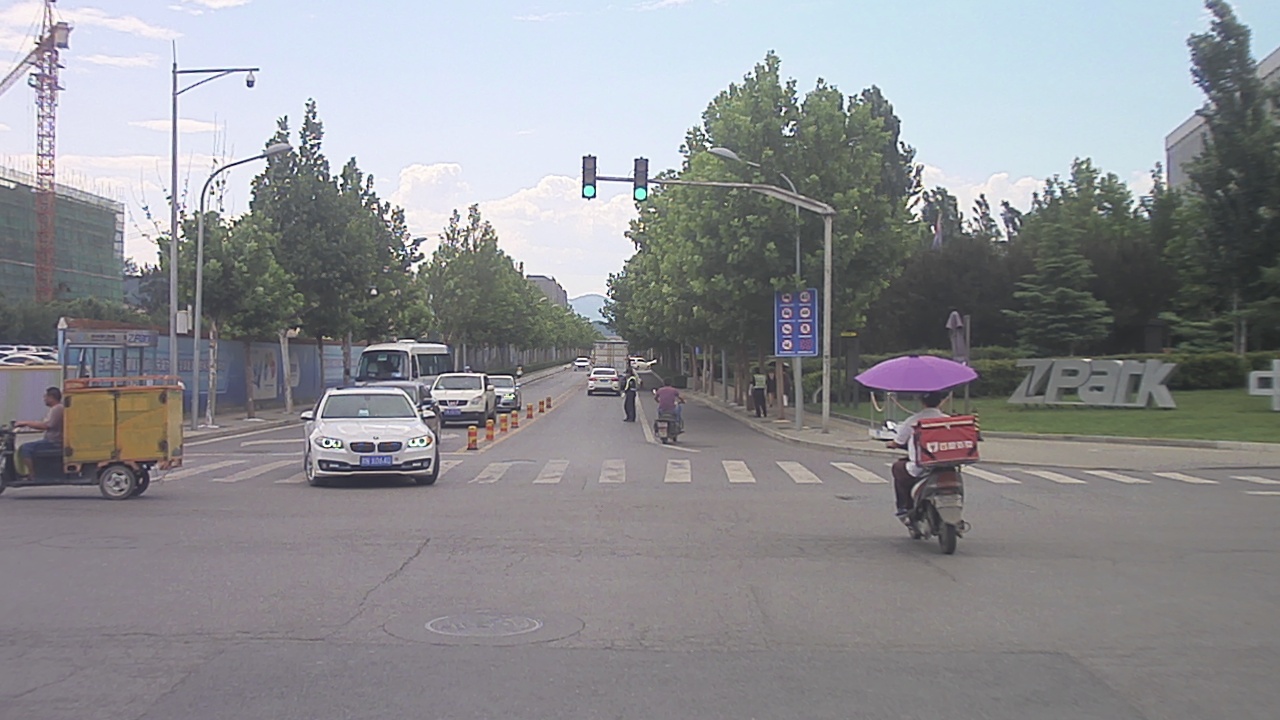}\hspace{-0.06 cm}
    \includegraphics[width=0.22\linewidth]{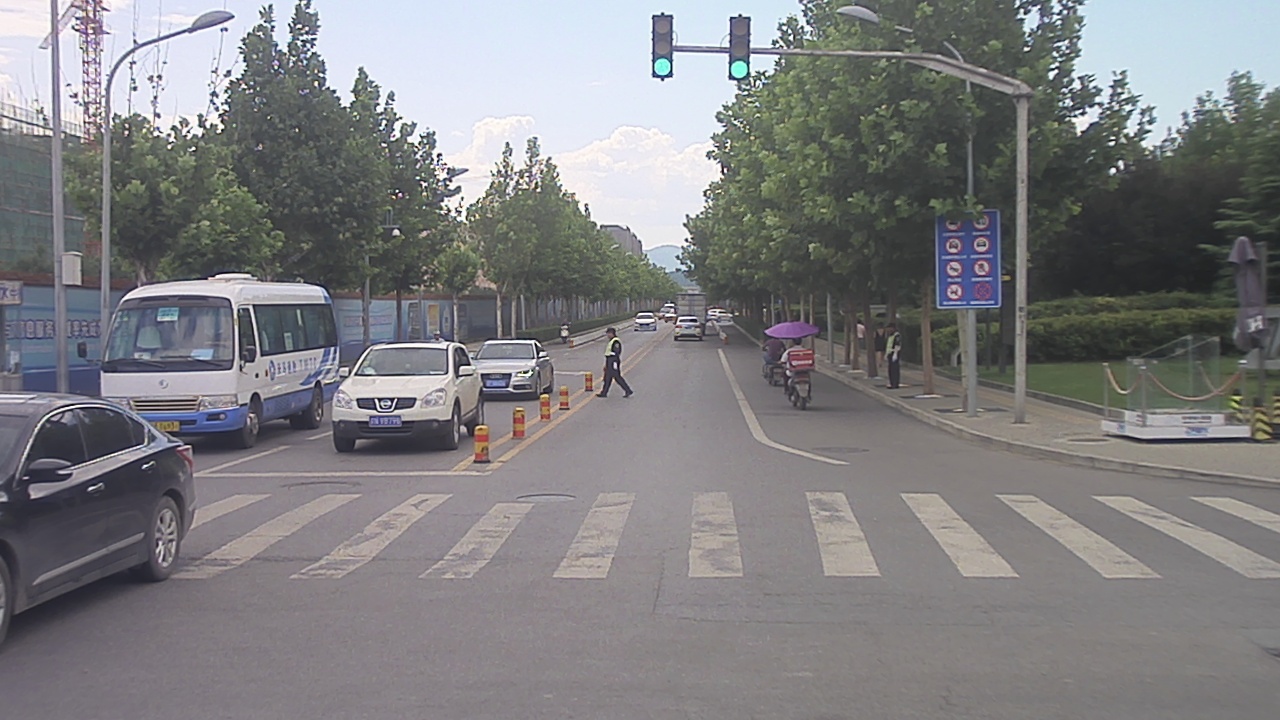}
    \end{minipage}

    \begin{minipage}{\textwidth}
    \centering
    \includegraphics[width=0.22\linewidth]{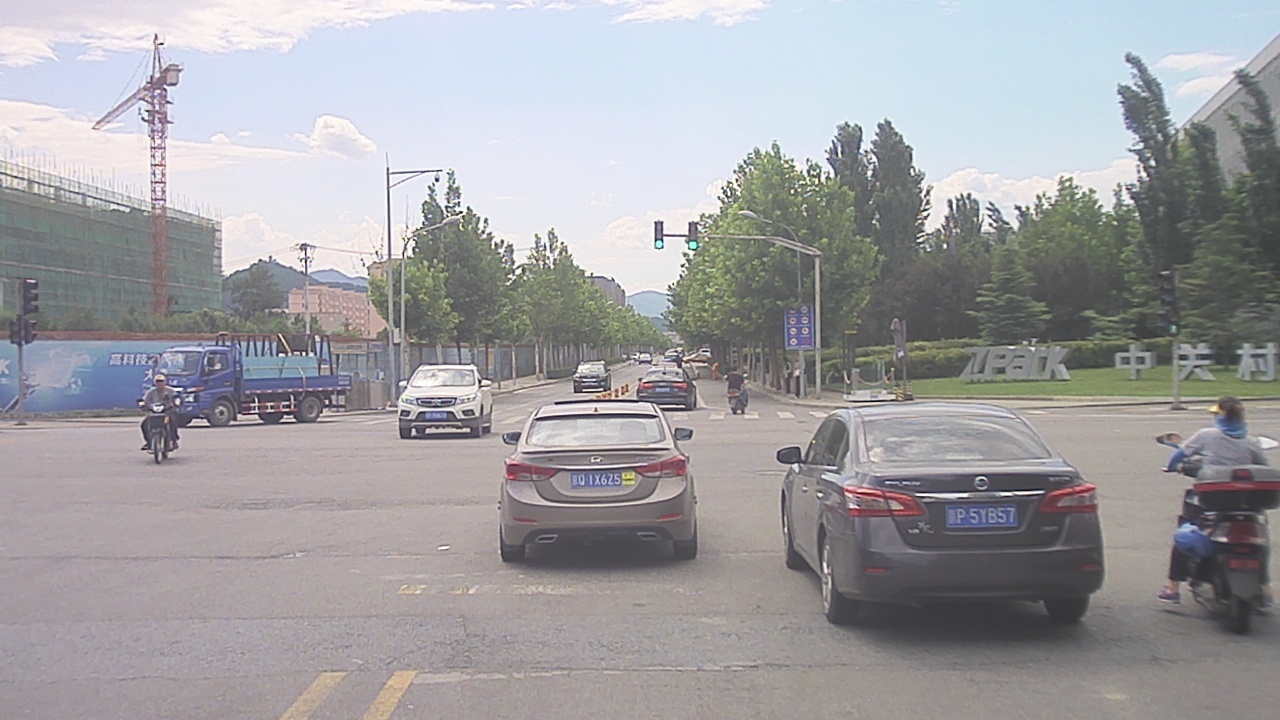}\hspace{-0.06 cm}
    \includegraphics[width=0.22\linewidth]{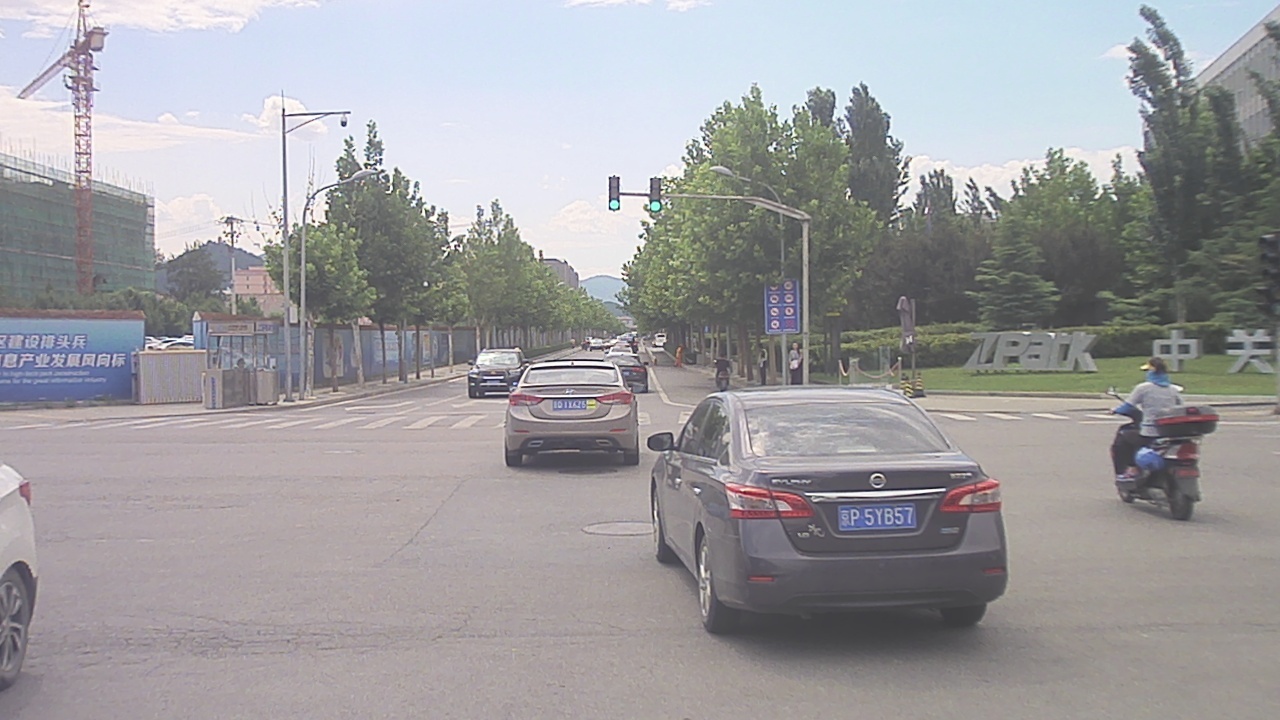}\hspace{-0.06 cm}
    \includegraphics[width=0.22\linewidth]{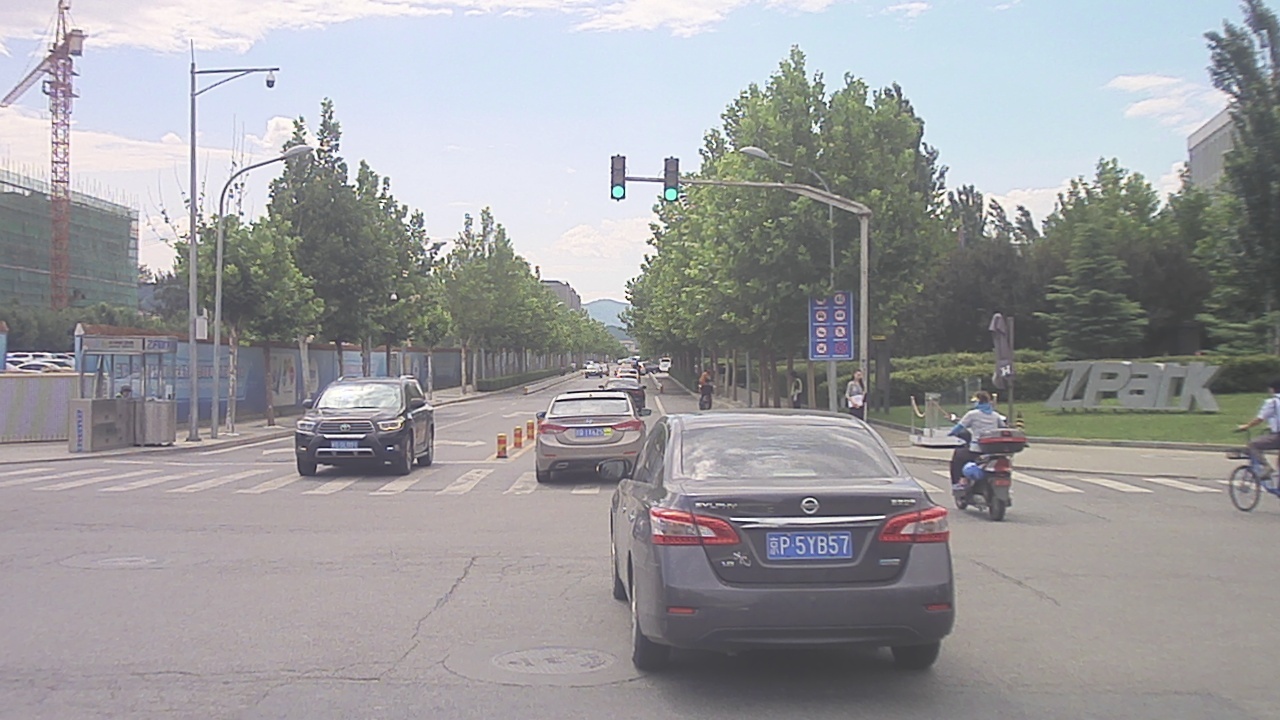}\hspace{-0.06 cm}
    \includegraphics[width=0.22\linewidth]{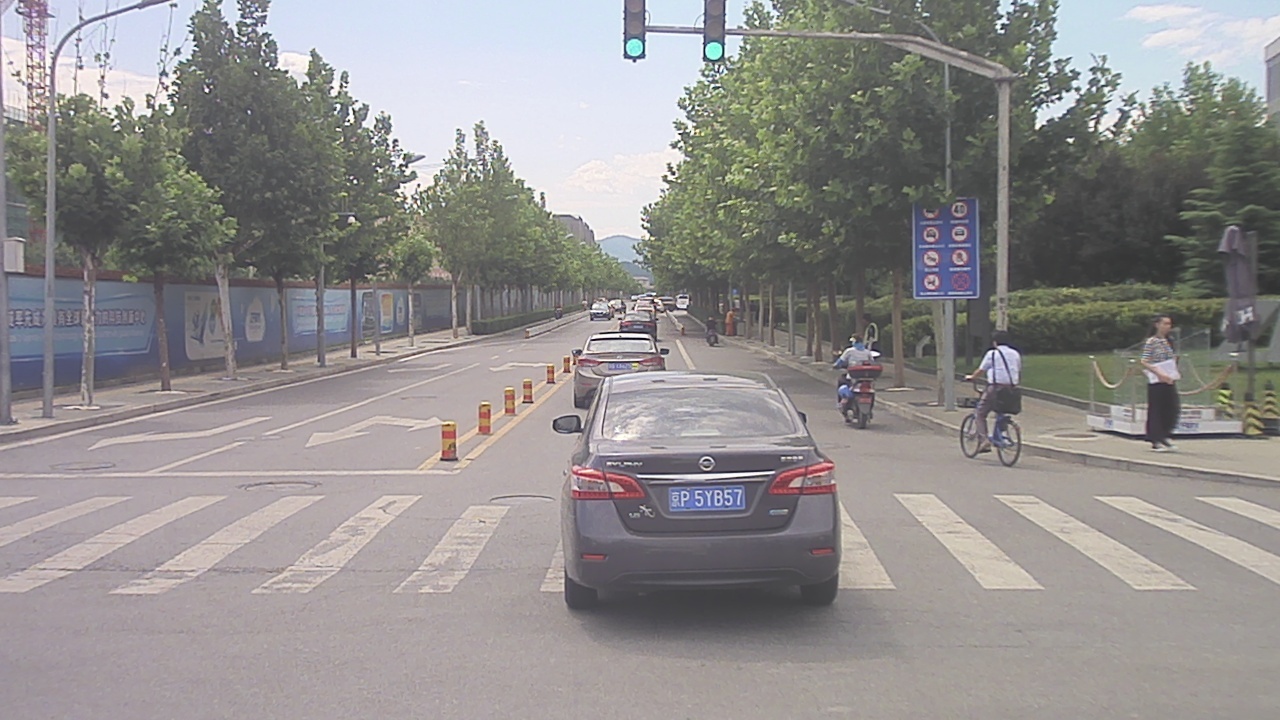}
    \end{minipage}

    \begin{minipage}{\textwidth}
    \centering
    \includegraphics[width=0.22\linewidth]{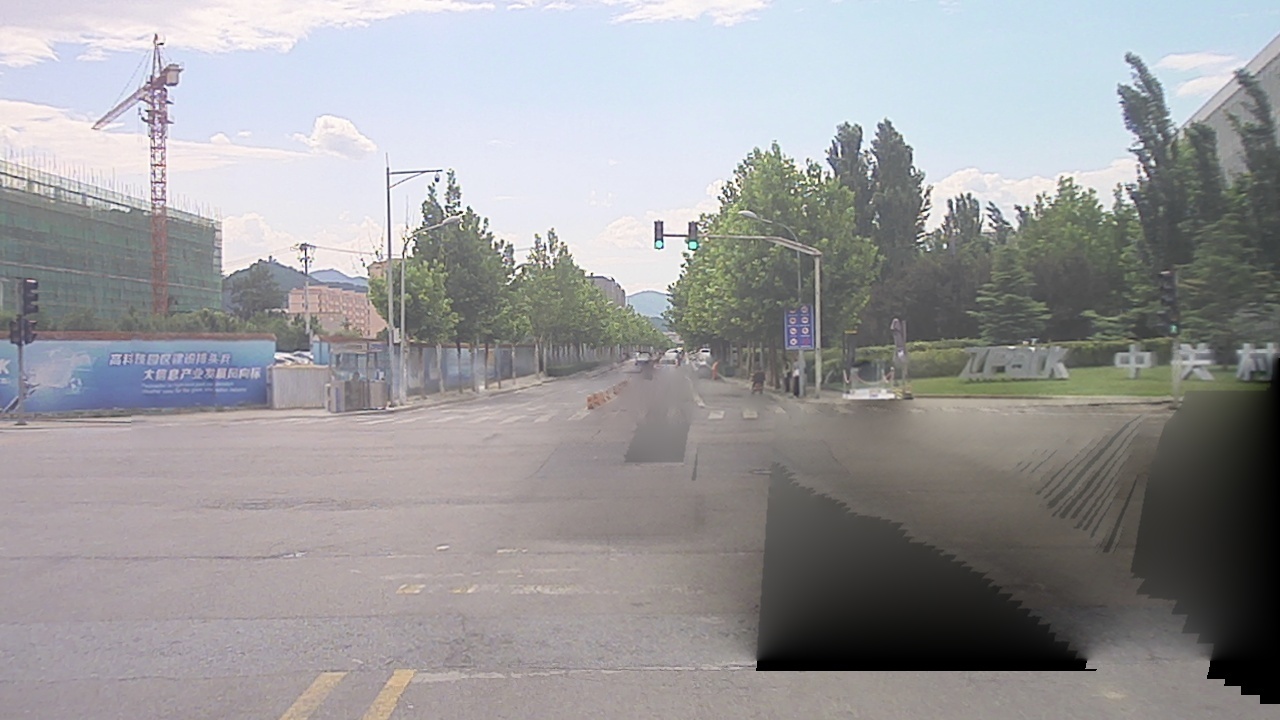}\hspace{-0.06 cm}
    \includegraphics[width=0.22\linewidth]{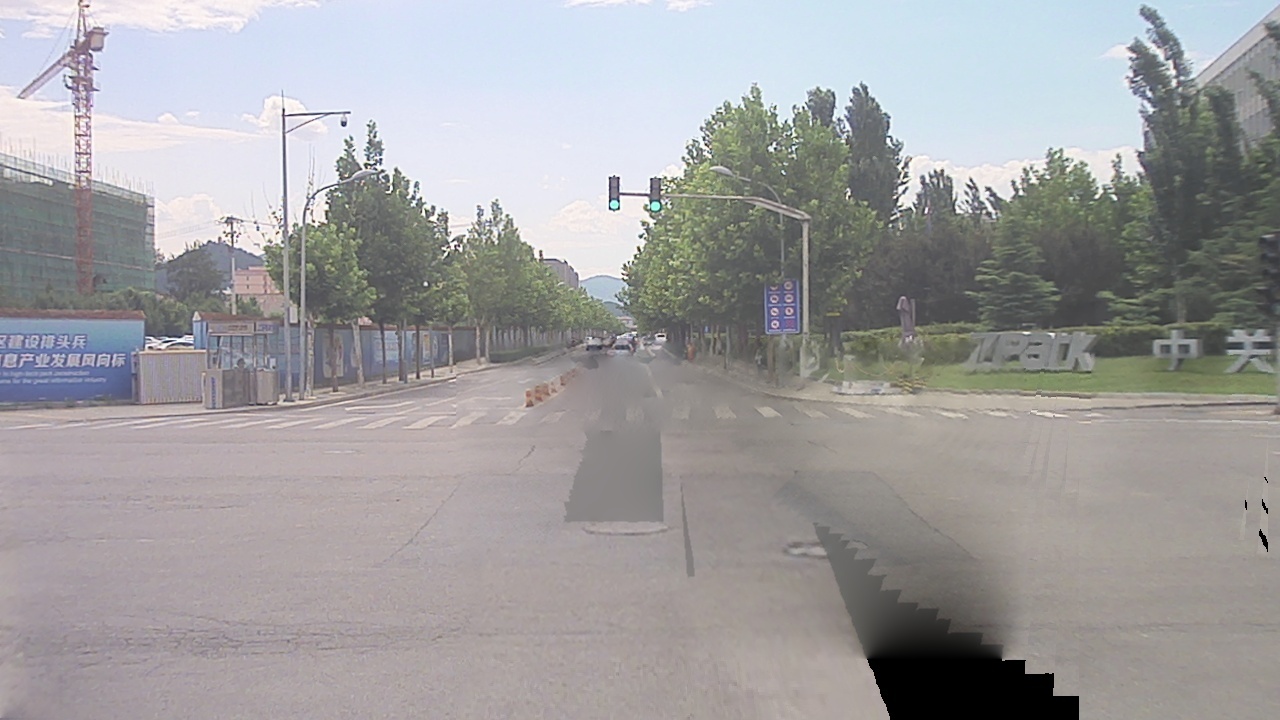}\hspace{-0.06 cm}
    \includegraphics[width=0.22\linewidth]{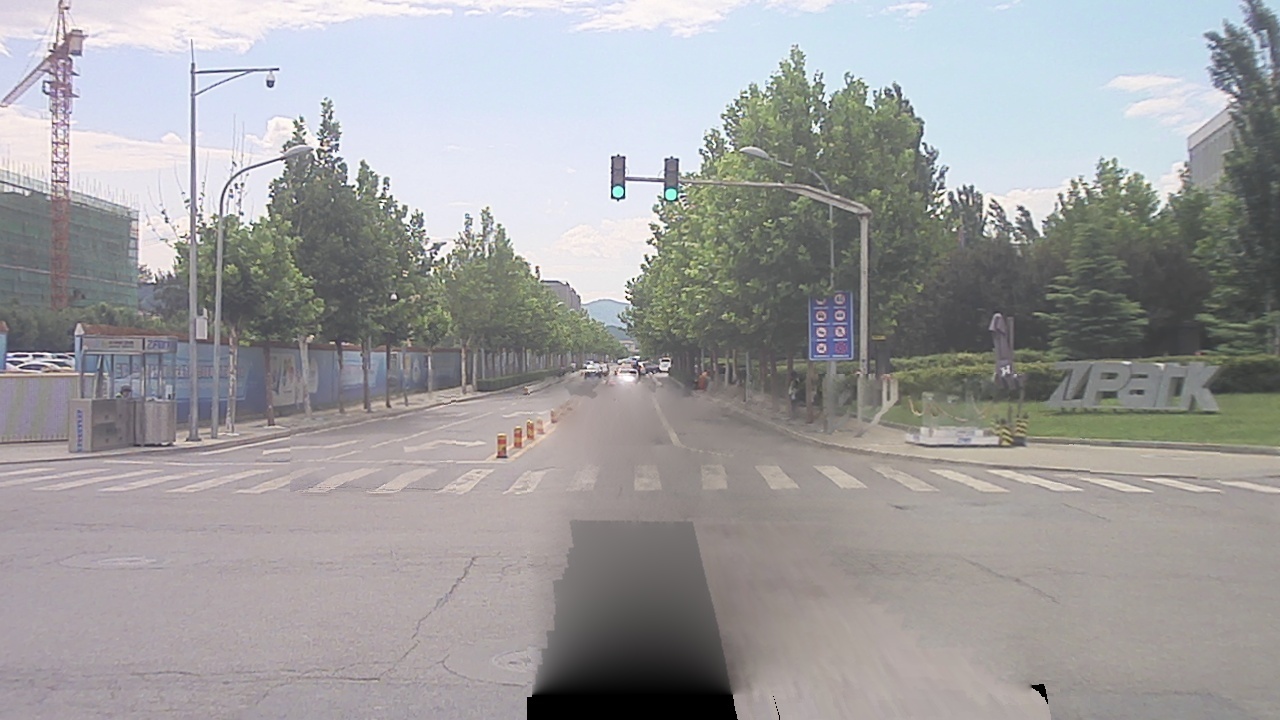}\hspace{-0.06 cm}
    \includegraphics[width=0.22\linewidth]{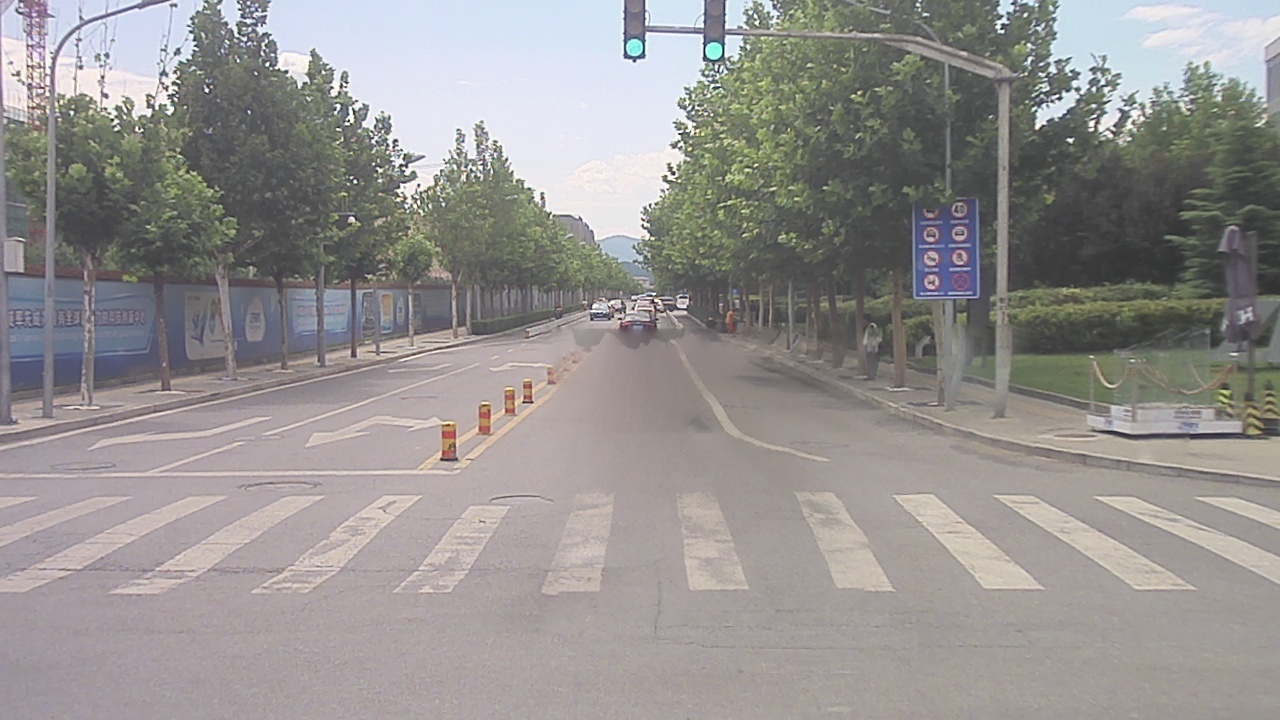}
    \end{minipage}

    \begin{minipage}{\textwidth}
    \centering
    \includegraphics[width=0.22\linewidth]{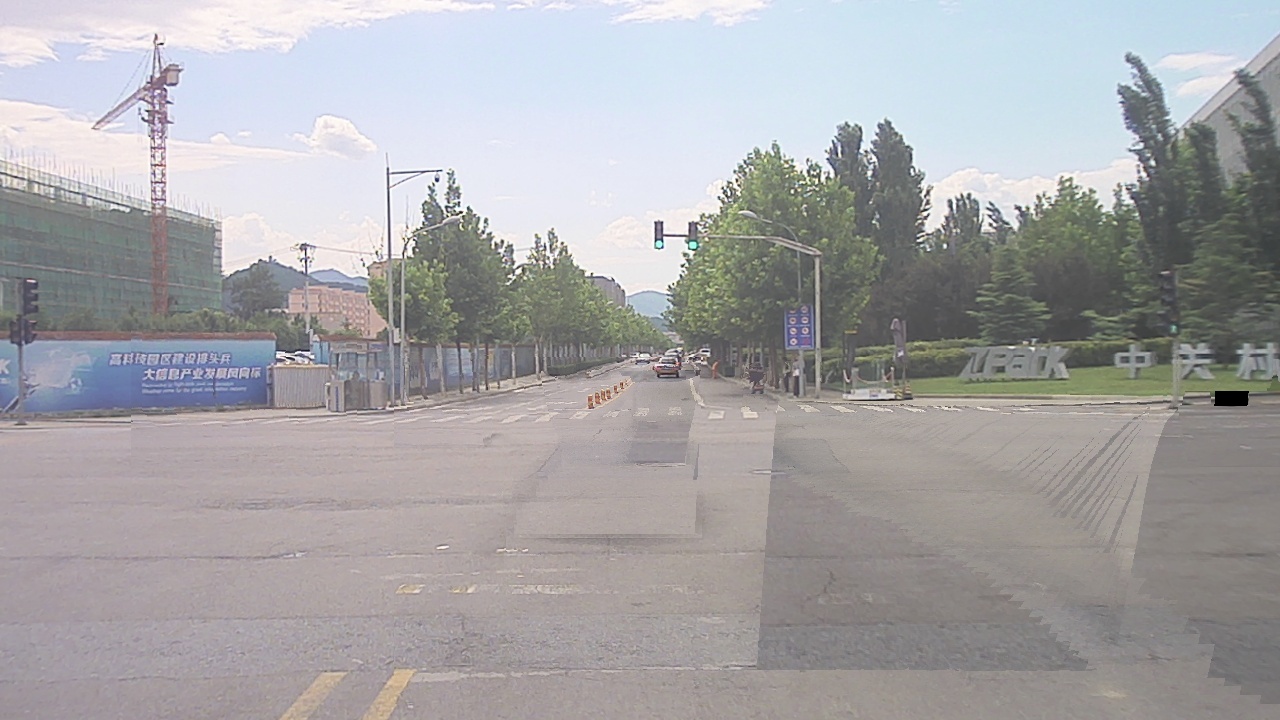}\hspace{-0.06 cm}
    \includegraphics[width=0.22\linewidth]{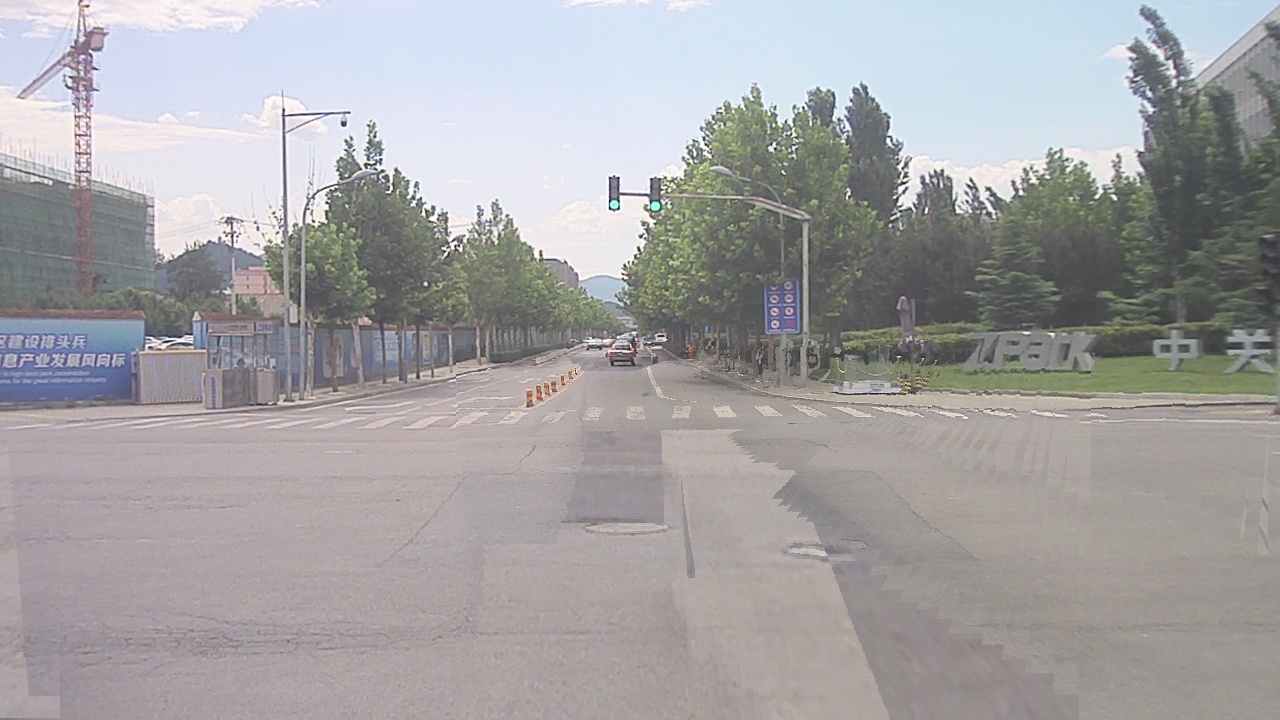}\hspace{-0.06 cm}
    \includegraphics[width=0.22\linewidth]{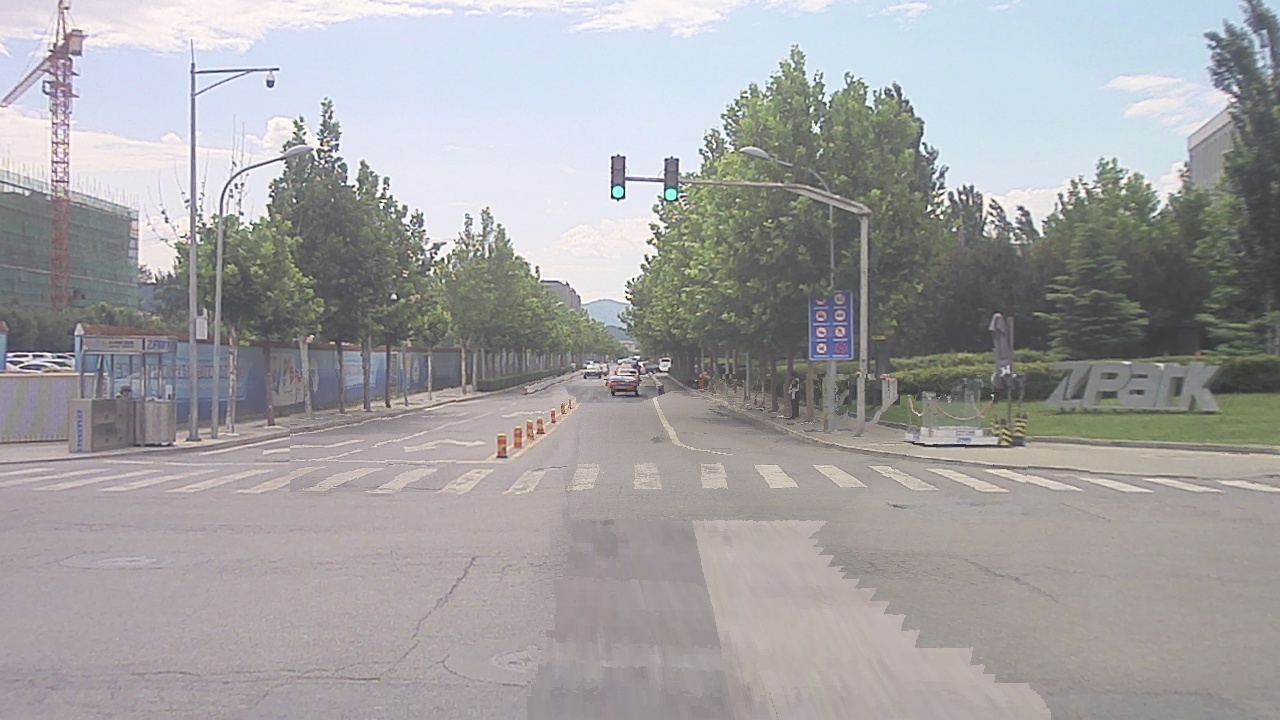}\hspace{-0.06 cm}
    \includegraphics[width=0.22\linewidth]{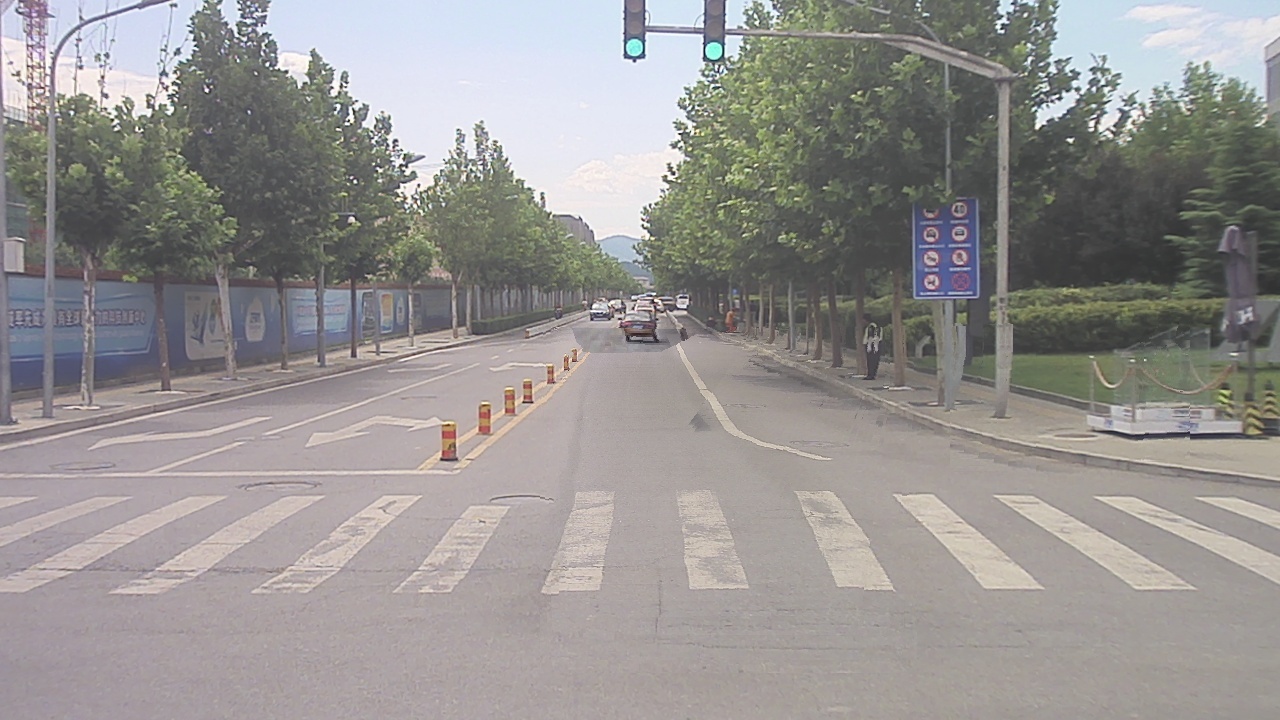}
    \end{minipage}

    \begin{minipage}{\textwidth}
    \centering
    \includegraphics[width=0.22\linewidth]{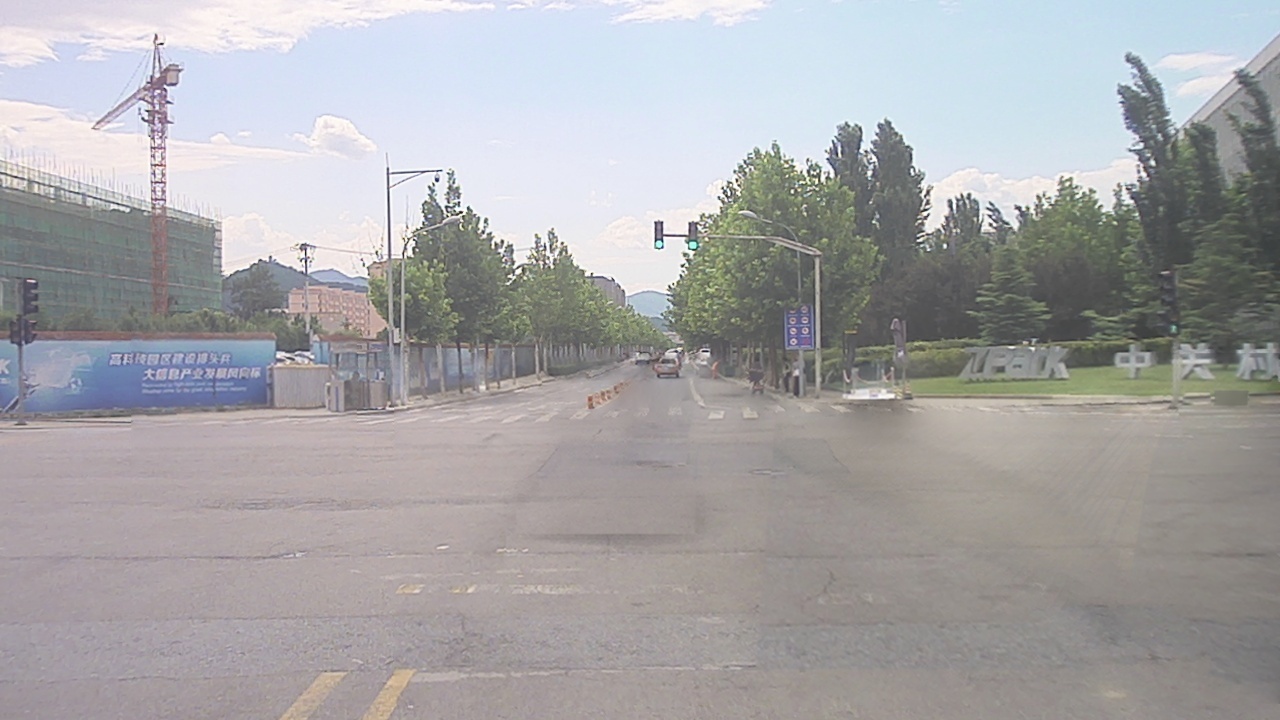}\hspace{-0.06 cm}
    \includegraphics[width=0.22\linewidth]{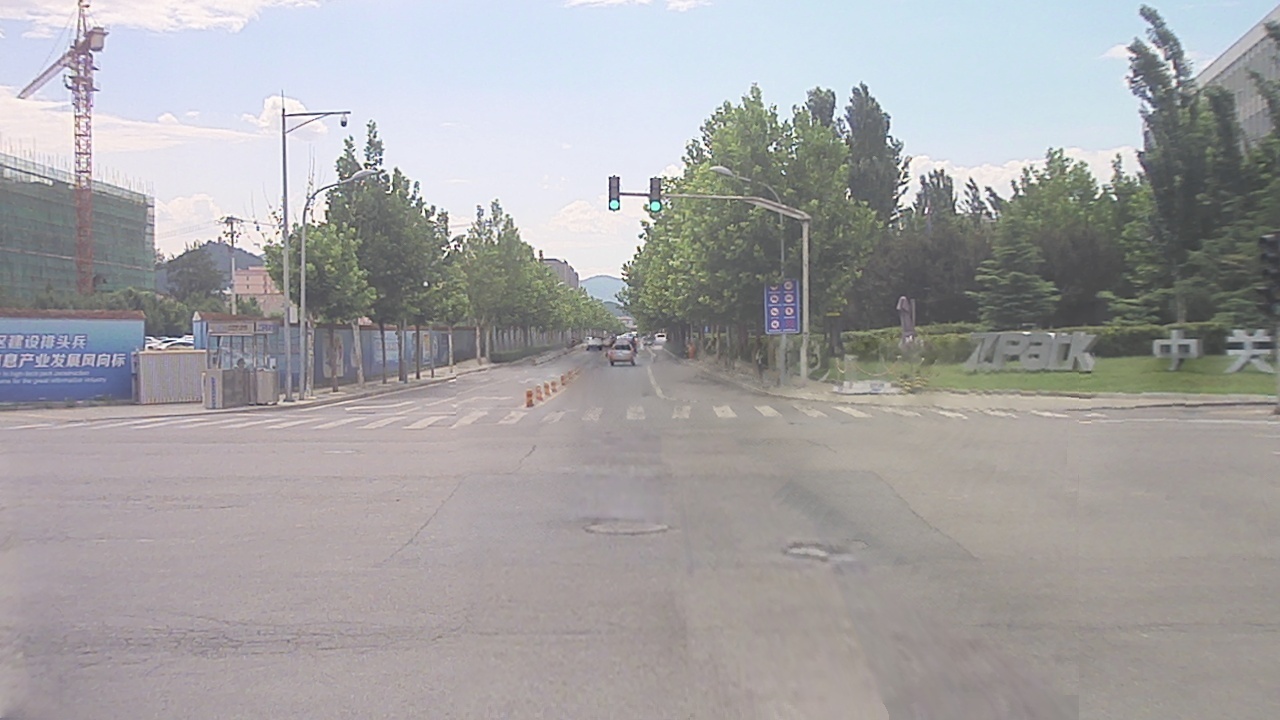}\hspace{-0.06 cm}
    \includegraphics[width=0.22\linewidth]{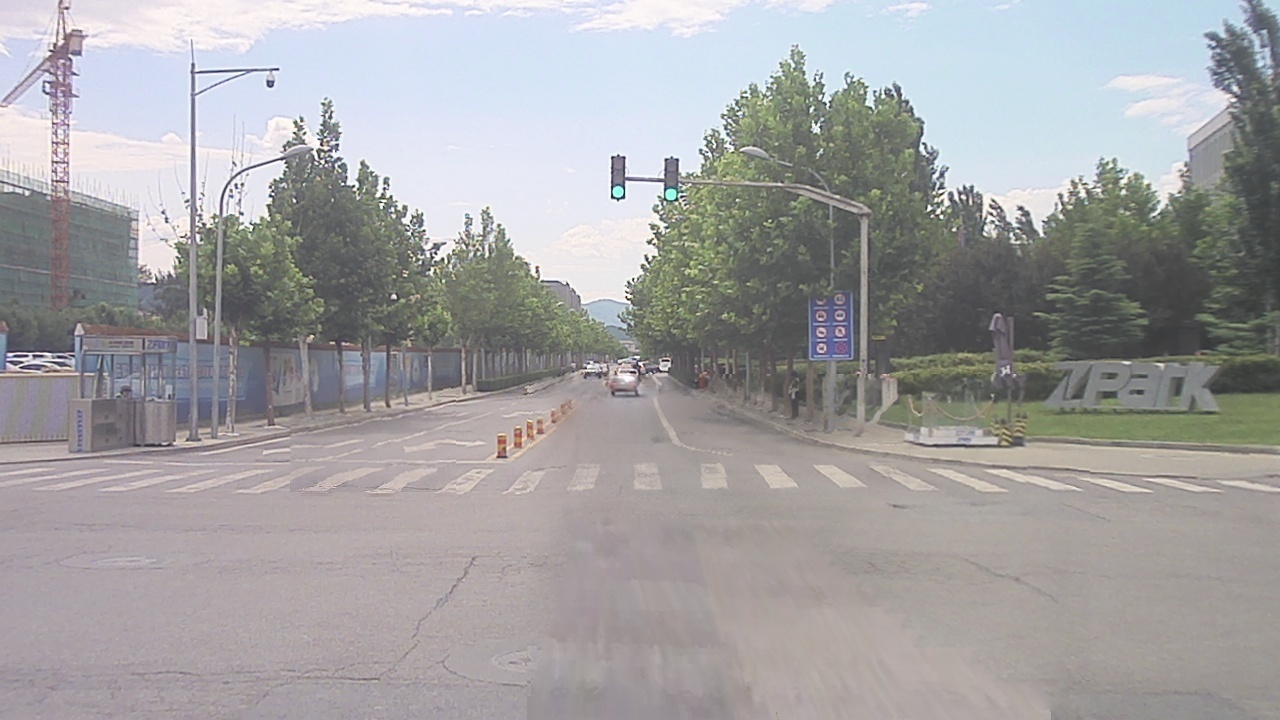}\hspace{-0.06 cm}
    \includegraphics[width=0.22\linewidth]{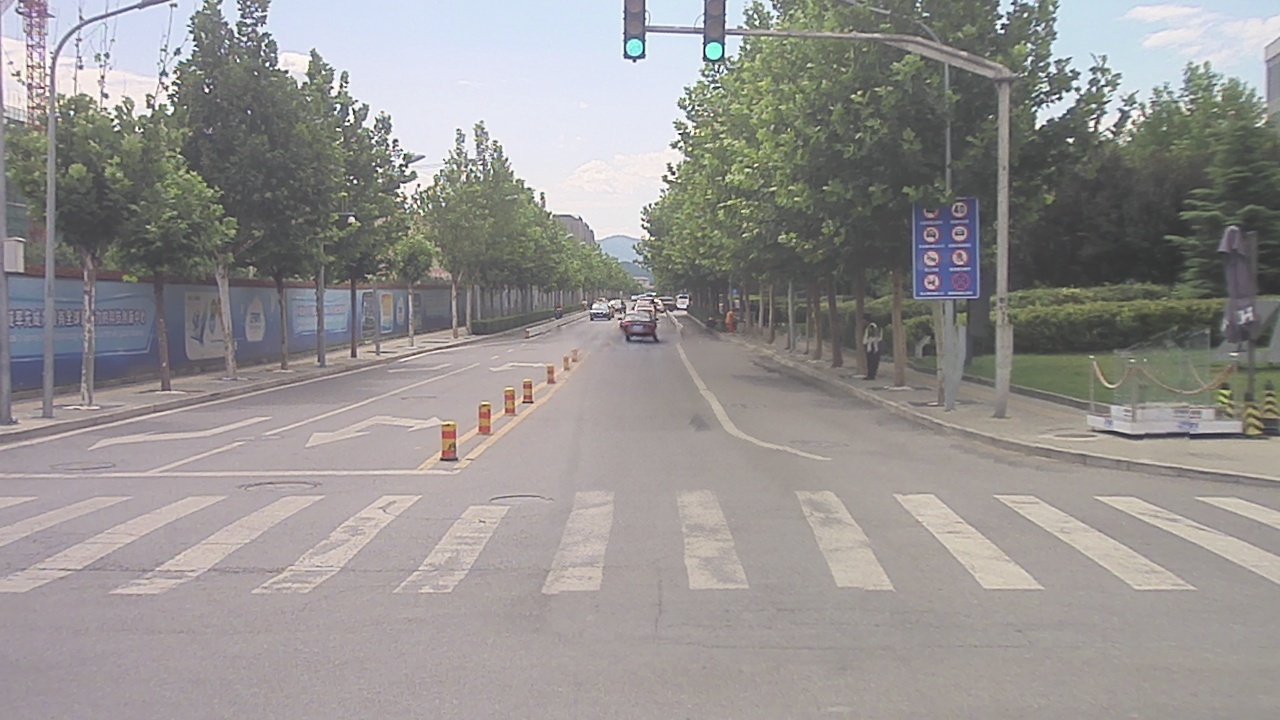}
    \end{minipage}

\caption{\textbf{1st row}: frames from video 1; \textbf{2nd row}: frames from video 2 captured on a different day; \textbf{3rd row}: results after Poisson color blending using video 2 only; \textbf{4th row}: direct inpainting results by fusing both videos; \textbf{5th row}: results after Poision color blending using both videos.}
\label{fig:fusion}
\end{figure*}

\textbf{Poisson Image Blending} Fig.~\ref{fig:fusion} shows the effectiveness of applying Poisson image blending. Visible seams are obvious at boundaries of pixels coming from different frames. This is because our capturing camera is working under auto exposure and auto white balance mode. A same object may have different color tones in different frames from the same video, not to mention videos captured on different days. Tab.~\ref{tab:colorblending} shows the quantitative results with and without Poisson color blending. It is clear that color blending indeed improves the results.

\begin{table}[]
% \parbox{.45\linewidth}{
\begin{minipage}[t]{.48\textwidth}
\centering
%\parbox{.45\linewidth}
{
\begin{tabular}{ l c c c c }
\hline
  Strategies   & MAE     & RMSE            & PSNR            & SSIM            \\ \hline
no blending     & 9.410          & 17.484          & 21.783          & 0.911          \\  
blending        &  \textbf{6.497}        &  \textbf{13.009}          &  \textbf{22.312}          &  \textbf{0.917}         \\ \hline
\end{tabular}
\caption{Ablation study on Poisson color blending, where the best results are highlighted in bold. To be clear, the values of ``MAE'' and ``RMSE'' are the lower the better while the values of ``PSNR'' and ``SSIM'' are the higher the better.}
\label{tab:colorblending}
}
\end{minipage}%
\hspace{0.4cm}%
\begin{minipage}[t]{.48\textwidth}

% \hfill
% \parbox{.45\linewidth}{
\centering
%\parbox{.45\linewidth}
{
\begin{tabular}{ l c c c c }
\hline
 Strategies & MAE             & RMSE            & PSNR            & SSIM            \\ \hline
no fusion           & 10.427          & 14.967          & 20.941          & 0.879          \\  
fusion              & \textbf{6.059}          &  \textbf{8.333}          &  \textbf{21.195}          &  \textbf{0.882}          \\ \hline
\end{tabular}
\caption{Ablation study on multiple video fusion, where the best results are highlighted in bold. To be clear, the values of ``MAE'' and ``RMSE'' are the lower the better while the values of ``PSNR'' and ``SSIM'' are the higher the better.}
\label{tab:fusion}
}
\end{minipage}
\end{table}

% \begin{table}[]
% \centering
% %\parbox{.45\linewidth}
% {
% \begin{tabular}{ l c c c c }
% \hline
%   Different Strategies   & MAE     & RMSE            & PSNR            & SSIM            \\ \hline
% Without blending     & 9.410          & 17.484          & 21.783          & 0.911          \\  
% With blending        &  \textbf{6.497}        &  \textbf{13.009}          &  \textbf{22.312}          &  \textbf{0.917}         \\ \hline
% \end{tabular}
% \caption{Ablation study on Poisson color blending, where the best results are highlighted in bold. To be clear, the values of ``MAE'' and ``RMSE'' are the lower the better while the values of ``PSNR'' and ``SSIM'' are the higher the better.}
% \label{tab:colorblending}
% }
% \end{table}

\textbf{Video Fusion}. Fig.~\ref{fig:fusion} shows fusion of 2 videos. 1st row shows four frames from a video and the 2nd row shows 4 frames from another video captured on a different day at the same traffic intersection. Here, our goal is to inpaint those foreground objects in the 2nd video. 3rd row shows output using video 2 only, where exists large blank regions. That is because front vehicles keep blocking certain areas during the entire capture time. It is clear that Poisson image blending is not capable of completing large blank holes. 4th row shows BP output after we fuse the 1st video into the 2nd one, where the blank holes are all gone. 5th row shows the final results after color blending and optical flow temporal smoothing. Tab.~\ref{tab:fusion} demonstrates effectiveness of video fusion.

% \begin{table}
% \centering
% %\parbox{.45\linewidth}
% {
% \begin{tabular}{ l c c c c }
% \hline
%  Different Strategies & MAE             & RMSE            & PSNR            & SSIM            \\ \hline
% Without fusion           & 10.427          & 14.967          & 20.941          & 0.879          \\  
% With fusion              & \textbf{6.059}          &  \textbf{8.333}          &  \textbf{21.195}          &  \textbf{0.882}          \\ \hline
% \end{tabular}
% \caption{Ablation study on multiple video fusion, where the best results are highlighted in bold. To be clear, the values of ``MAE'' and ``RMSE'' are the lower the better while the values of ``PSNR'' and ``SSIM'' are the higher the better.}
% \label{tab:fusion}
% }
% \end{table}

The fusion of multiple videos for inpainting demonstrates another advantage of our proposed approach. For those existing video inpainting works, they haven't address the issue of long-time occlusion, neither did they proposed to fuse multiple videos for inpainting purpose. Please checkout video demos here: \url{https://youtu.be/iOIxdQIzjQs} .

%% file: 5-con.tex
\section{Conclusion}
In this paper, we propose an automatic video inpainting algorithm that removes object from videos and synthesizes missing regions with the guidance of depth. It outperforms existing state-of-the-art inpainting methods on our inpainting dataset by preserving accurate texture details. The experiments indicate that our approach could  reconstruct cleaner and better background images, especially in the challenging scenarios with long time occlusion scenes. Furthermore, our method may be generalized to any videos as long as depth exists, in contrast to those deep learning-based approaches whose success heavily depend on comprehensiveness and resemblance of training dataset. 

%Currently, we only use forward-facing color image sensor. In the future, we plan to use 360 degree panoramic camera to improve our inpainting result. This way, we can largely alleviate the occlusion issue by using pixels of side and back views. 

% \textbf{fuse data from different capture} As illustrated above, our algorithm could leave some pixels not inpainted due to occlusion. Currently, we solve this by employing other inpainting method in the post processing. A better way to address this issue would be capturing another video of the same scene, where the occluded parts become visible. It is straightforward to register newly captured frames into existing 3D map using LOAM~\cite{Zhang-2014-7903}. Once new frames are registered and merged into existing 3D map, inpainting is performed exactly the same way.

% \textbf{color unifying} As can be seen from some of our results, pixels from different frames have slightly different color intensity and temperature, resulting in visible color seams in a few inpainting regions. If we introduce panoramic camera or even data captured from different time, we are facing bigger issue. In order to overcome this, we should leverage techniques of Poisson image editing~\cite{Perez:2003:PIE:1201775.882269} or pyramid image blending~\cite{1095851} to fuse pixels of different colors.